\documentclass{article}

\usepackage{times}
\usepackage[utf8]{inputenc} %
\usepackage[T1]{fontenc}    %
\usepackage[hidelinks]{hyperref}       %
\usepackage{natbib}
\usepackage{url}            %
\usepackage{booktabs}       %
\usepackage{amsfonts}       %
\usepackage{nicefrac}       %
\usepackage{microtype}      %
\usepackage{enumitem}

\usepackage{graphicx}
\usepackage{sidecap}
\usepackage{diagbox}
\usepackage[small]{caption}
\usepackage{subcaption}
\usepackage{comment} 
\usepackage{array}
\usepackage[export]{adjustbox}

\usepackage{float} %
\usepackage{wrapfig}
\usepackage{mathtools}
\usepackage{xspace}
\usepackage{bm} %

\newcommand{\rightcomment}[1]{\(\triangleright\) {\small \it #1}}  %
\newcommand{\eqcomment}[1]{\addtocounter{equation}{1}\tag*{\rightcomment{#1}\quad(\theequation)}}  %
\usepackage{suffix}
\WithSuffix\newcommand\eqcomment*[1]{\tag*{\rightcomment{#1}}}  %

\usepackage{algorithm}
\usepackage[noend]{algpseudocode}

\renewcommand\algorithmicdo{:}
\renewcommand\algorithmicthen{:}
\algnewcommand{\IfThen}[2]{\State \algorithmicif\ #1\ \algorithmicthen\ #2}
\algnewcommand{\IfThenElse}[3]{\State \algorithmicif\ #1\ \algorithmicthen\ #2\ \algorithmicelse\ #3}
\algrenewcommand{\algorithmiccomment}[1]{\hfill \rightcomment{#1}}
\algnewcommand{\LineComment}[1]{\State \rightcomment{#1}}
\algnewcommand{\LinesComment}[1]{\State \rightcomment{\parbox[t]{\linewidth-\leftmargin-\widthof{\(\triangleright\) }}{#1}}\smallskip}

\algnewcommand\algorithmicinput{{\bfseries Input:}}
\algnewcommand\INPUT{\item[\algorithmicinput]}
\algnewcommand\algorithmicoutput{{\bfseries Output:}}
\algnewcommand\OUTPUT{\item[\algorithmicoutput]}
\makeatletter
\newcounter{algorithmicH}
\let\oldalgorithmic\algorithmic
\renewcommand{\algorithmic}{%
  \stepcounter{algorithmicH}
  \oldalgorithmic}
\renewcommand{\theHALG@line}{ALG@line.\thealgorithmicH.\arabic{ALG@line}}
\makeatother
\makeatletter
\newcommand{\algmargin}{\the\ALG@thistlm}
\makeatother
\algnewcommand{\Statepar}[1]{\State\parbox[t]{\dimexpr\linewidth-\algmargin}{\strut #1\strut}}

\usepackage{mdframed} %

\usepackage[usenames,dvipsnames]{xcolor}
\usepackage[textwidth=1.5cm]{todonotes} %
\usepackage{pgfplots}

\newcommand{\para}[1]{\noindent \textbf{#1}}
\newcommand{\cutforspace}[1]{}

\newif\ifhidecomments

\ifhidecomments
    \newcommand{\chenhao}[1]{}
    \newcommand{\hongyuan}[1]{}
    \newcommand{\rosa}[1]{}
    \newcommand{\haokun}[1]{}
    \newcommand{\tejes}[1]{}
\else
    \newcommand{\chenhao}[1]{\textcolor{blue}{[\textsc{Chenhao}: #1]}}
    \newcommand{\hongyuan}[1]{\textcolor{green!70!blue}{[\textsc{Hongyuan}: #1]}}
    \newcommand{\rosa}[1]{\textcolor{magenta!80!brown}{[\textsc{Rosa}: #1]}}
    \newcommand{\haokun}[1]{\textcolor{orange!60!brown}{[\textsc{Haokun}: #1]}}
    \newcommand{\tejes}[1]{\textcolor{red!80!blue}{[\textsc{Tejes}: #1]}}
\fi

\usepackage{listings}

\usepackage{parcolumns}
\lstdefinestyle{datalogstyle}{
        basicstyle={\tt \scriptsize},  %
	xleftmargin={6pt},
        xrightmargin={6pt},
        columns=flexible,
        breakindent=0pt,
        breaklines=true, 
	frame=tb,
	stepnumber=1,
	firstnumber=1,
	numberfirstline=true,
	tabsize=2,
	extendedchars=true,
	breaklines=true,
	columns=fullflexible,
	keepspaces=true,
	escapeinside={@}{@},
	firstnumber=last,
	captionpos=b, 
	commentstyle=\color{black!65},
	numberstyle=\tiny\color{black!65},
	stringstyle=\color{codepurple},
	breakatwhitespace=false, 
	keepspaces=true,              
        mathescape=true, 
	numbersep=5pt,                  
	showspaces=false,                
	showstringspaces=false,
	showtabs=false,
	aboveskip={0.8\baselineskip},
	belowskip={0.2\baselineskip},
}
\definecolor{aigold}{RGB}{244,210, 1} 
\definecolor{aigreen}{RGB}{213, 245, 227}

\definecolor{humanpurple}{RGB}{235, 222, 240} 
\definecolor{mypurple}{RGB}{147,112,219} 
\definecolor{myorange}{RGB}{255,165,0} 
\definecolor{commentgray}{RGB}{86, 101, 115}
\definecolor{mygray}{RGB}{169,169,169}
\definecolor{aired}{RGB}{255,180,181}

\usepackage[noabbrev,capitalize]{cleveref} %
\crefname{equation}{equation}{equations}   %
\crefname{footnote}{footnote}{footnotes}   
\crefname{listing}{Example}{Examples}
\crefname{assumption}{assumption}{assumptions}
\crefname{line}{line}{lines}   %
\crefname{section}{\S}{\S\S}

\newcommand{\calS}{\mathcal{S}}

\newcommand{\mixtral}{Mixtral\xspace}
\newcommand{\claude}{Claude-2.1\xspace}
\newcommand{\gpt}{GPT-3.5-turbo\xspace}
\newcommand{\roberta}{RoBERTa\xspace}
\newcommand{\llama}{Llama-2-7B\xspace}
\newcommand{\finetune}{the fine-tuned models\xspace}
\newcommand{\deceptive}{\textsc{Deceptive reviews}\xspace}
\newcommand{\headline}{\textsc{Headline popularity}\xspace}
\newcommand{\retweet}{\textsc{Tweet popularity}\xspace}
\newcommand{\shoe}{\textsc{Shoe sales}\xspace}
\newcommand{\ours}{{\bf HypoGeniC}\xspace}

\definecolor{pastelgreen}{RGB}{50,205,50}
\newcommand{\increase}{\textcolor{pastelgreen}{\bm{$\uparrow$}}}
\newcommand{\decrease}{\textcolor{red}{\bm{$\downarrow$}}}

\newcommand{\novel}{\includegraphics[width=0.8cm,valign=c]{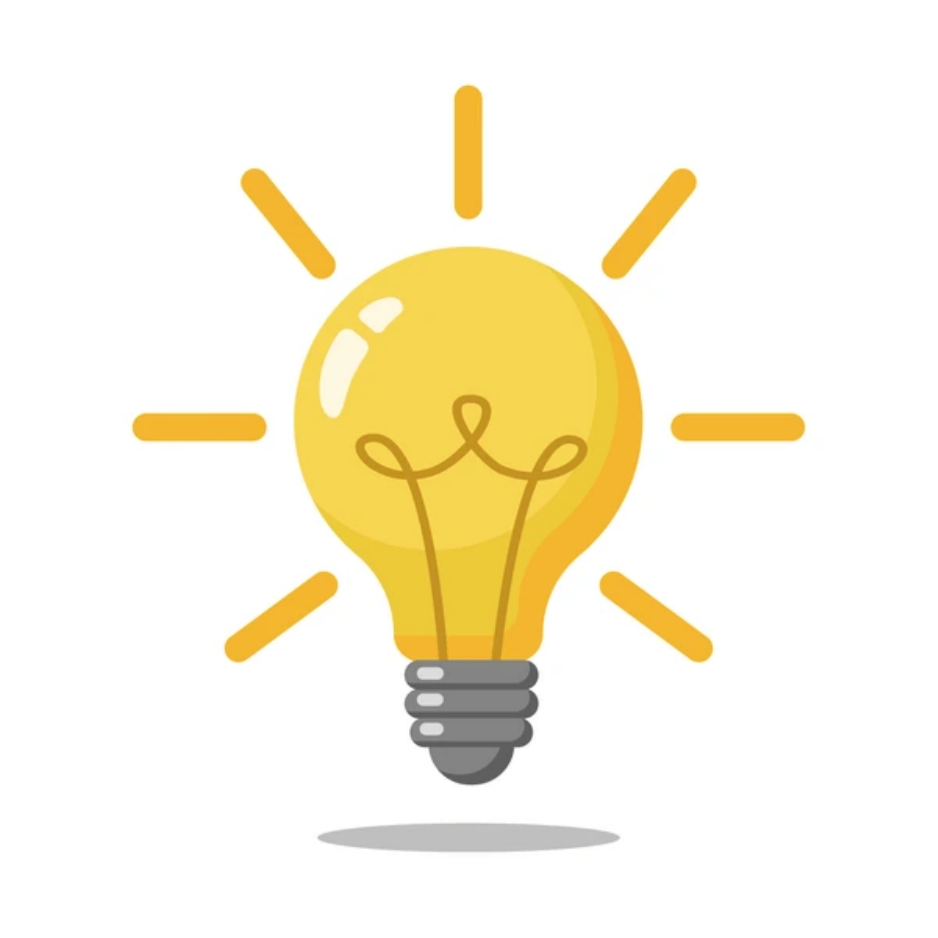}}
\title{Hypothesis Generation with Large Language Models}

\usepackage[final]{acl}

\author{Yangqiaoyu Zhou$^\clubsuit$, Haokun Liu$^\clubsuit$,  Tejes Srivastava$^\clubsuit$\\
\textbf{Hongyuan Mei}$^\dag$ \& \textbf{Chenhao Tan}$^\clubsuit$  \\
Department of Computer Science\\
University of Chicago$^\clubsuit$, Toyota Technological Institute at Chicago$^\dag$\\
Chicago, IL 60637, USA \\
\texttt{\{zhouy1,haokunliu,tejess,chenhao\}@uchicago.edu,hongyuan@ttic.edu} \\
}

\begin{document}

\maketitle
\begin{abstract}

Effective generation of novel hypotheses is instrumental to scientific progress.
So far, researchers have been the main powerhouse behind hypothesis generation by painstaking data analysis and thinking (also known as the Eureka moment).
In this paper, we examine the potential of large language models (LLMs) to generate hypotheses.
We focus on hypothesis generation based on data (i.e., labeled examples).
To enable LLMs to handle 
long contexts,
we generate initial hypotheses from a small number of examples
and then update them iteratively to improve the quality of hypotheses.
Inspired by multi-armed bandits, we design a reward function to inform the exploitation-exploration tradeoff in the update process.
Our algorithm is able to generate hypotheses that enable much better predictive performance than few-shot prompting in classification tasks, improving accuracy by 31.7\% on a synthetic dataset and by 13.9\%, 3.3\% and, 24.9\% on three real-world datasets.
We also outperform supervised learning by 12.1\% and 11.6\% on two challenging real-world datasets.
Furthermore, we find that the generated hypotheses not only corroborate human-verified theories but also uncover new insights for the tasks. 
\end{abstract}

\section{Introduction}
\label{sec:introduction}

Hypothesis generation drives scientific progress.
Mendel's hypothesis on allele pairs lays the foundation for modern genetics;
Einstein's hypothesis in general theory of relativity led to the prediction and subsequent confirmation of gravitational waves.
In the context of language modeling, the hypothesis on scaling law inspires recent progress in large language models (LLMs)~\citep{ScalingLawsforLLM}.
Despite the importance of hypothesis generation, as \citet{ludwig+mullainathan:2024} point out, science has been curiously asymmetric.
While many scientific publications present extensive formal and empirical evaluation of hypotheses, the generation of hypotheses happens off-stage by researchers. 
In order to generate novel hypotheses, researchers may read literature, analyze data, pick the brain of each other, and even ``hallucinate'' (see Kekul\'e's discovery of the structure of the benzene molecule~\citep{rothenberg1995creative}).
Given the rise of large language models~\citep{brown2020language,anthropic2023claude2,openai2023gpt4}, we examine their potential of providing much needed assistance in hypothesis generation in this work.

In particular, we focus on hypothesis generation based on data, a common approach in empirical sciences.
Our main question is how we can enable LLMs to generate hypotheses of high-quality.
While one can easily prompt LLMs to generate hypotheses, LLMs may not be able to effectively leverage the input examples in a single long prompt.
Moreover, it is important to have measures of quality in the generation process so that we can filter bad hypotheses and come up with better ones.
These two observations motivate us to start with a setup analogous to supervised learning.
We can iteratively prompt an LLM to generate hypotheses based on the training examples and use training accuracy as a measure of quality to guide the generation process.
Conveniently, we can also evaluate the quality of the final generated hypotheses with their performance on held-out examples, similar to supervised learning.

\begin{figure*}[t]
    \centering
    \includegraphics[width=0.95\textwidth]{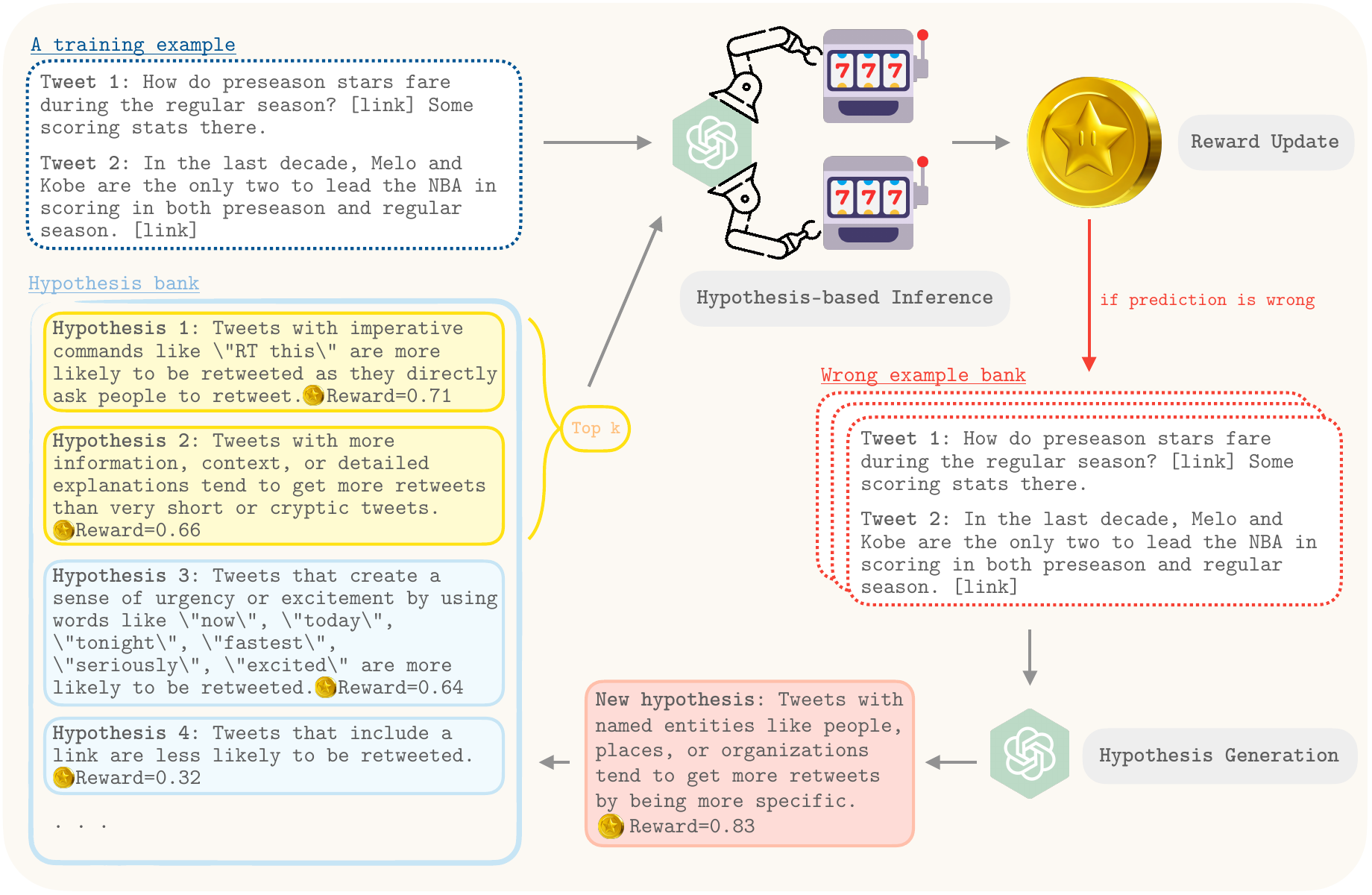}
    \caption{
    Illustration of \ours.
    During update stage, we evaluate the top $k$ hypotheses on each new training example and update the reward based on the prediction correctness. If the number of hypotheses that got the example wrong exceeds a certain threshold, we add the example to a wrong example bank.
    The wrong example bank is then used to generate new hypotheses.
    }
    \label{fig:update}
\end{figure*}

To generate high-quality hypotheses with LLMs, we propose an algorithm 
inspired by the upper confidence bound algorithm in multi-armed bandits \citep{auer2002using} (\ours \footnote{We have publicly released the code and data for \ours at \url{https://github.com/ChicagoHAI/hypothesis-generation}.}, {\bf Hypo}thesis {\bf Gen}eration {\bf i}n {\bf C}ontext; see \cref{fig:update}). 
Given initial hypotheses generated from a small number of examples, we need to assess their quality and propose new hypotheses to address their deficiencies.
To navigate this exploration-exploitation tradeoff,
we introduce a 
reward function 
and evaluate the top $k$ hypotheses for each 
training example.
We maintain a wrong example bank to capture the gap in knowledge of the hypotheses pool, and
generate new hypotheses based on the wrong example bank to close the gap.

The generated hypotheses naturally enable an interpretable hypothesis-based classifier.
We propose a suite of inference strategies given a set of hypotheses.
We apply our method to one synthetic task where there is a single known valid hypothesis and three real-world tasks (\deceptive, \headline, and \retweet).
The real-world tasks focus on deception detection and message popularity prediction, which are known to be challenging even for humans~\citep{ott2011finding,salganik2006experimental}.
Our algorithm can recover the hypothesis in the synthetic task and also provide useful hypotheses for the real-world tasks.
In fact, our generated hypotheses 
consistently outperform few-shot in-context learning baselines across all four tasks (31.7\% in \shoe, 13.9\% in \deceptive, 3.3\% in \headline, and 24.9\% in \retweet).
The predictive performance matches and even outperforms oracle supervised learning with \roberta and \llama except in \deceptive.

It is important to emphasize that although the utility of hypotheses in assisting downstream classification serves as an indicator for LLMs' ability to generate hypotheses, \textbf{our goal is not to maximize the classification performance.}
Rather, our primary interest lies in the \textbf{quality of the hypotheses}.
Thus, it is critical for the hypotheses to be interpretable beyond the LLM used to produce the hypotheses.
We show that hypotheses generated by one LLM (e.g., \gpt) can be used to make accurate inference by another LLM (e.g., \mixtral).
On an out-of-distribution dataset for \deceptive, we can even outperform the oracle fine-tuned \roberta. 
Such cross generalization provides strong evidence that we are able to generate hypotheses of high quality.
Furthermore, 
through a qualitative analysis, \textbf{our generated hypotheses
not only confirm theories from existing literature but also provide new insights about the task.}
For instance, one novel hypothesis is that ``reviews that mention personal experiences or special occasions, such as birthdays, anniversaries, or weddings, are more likely to be truthful''.
We encourage future research on deception detection to explore these novel hypotheses.

To summarize, we make the following contributions:
\begin{itemize}%
\item We propose a novel computational framework for generating and evaluating hypotheses with LLMs.
\item Our generated hypotheses enable interpretable hypothesis-based classifiers that outperform in-context learning and even supervised learning for one synthetic and three real-world datasets. These hypotheses are also robust across different LLMs and out-of-distribution datasets.
\item Our generated hypotheses corroborate existing findings while also providing new insights for the tasks.

\end{itemize}

\section{Method}
\label{sec:method}

We begin with a description of the problem formulation.
Given a set $\calS=\{(x_1, y_1), \ldots, (x_n, y_n)\}$ where $x_i$ is an example and $y_i$ is the corresponding label, the goal is to learn a set of hypotheses $\mathcal{H}=\{h_1, ..., h_m\}$ that describe theories of relationships 
between $x$ and $y$.
To this end, we prompt an LLM to summarize demonstration examples into high-level hypotheses (\cref{sec:hypotheses-generation}).
Then, during inference, the LLM makes inference based on the generated hypothesis (\cref{sec:hypothesis-based-inference}).

\subsection{Hypothesis Generation}
\label{sec:hypotheses-generation}

Our hypothesis generation algorithm (\cref{alg:hyp_gen}) is inspired by the upper confidence bound (UCB) algorithm \citep{auer2002using}.
Given a set of initial examples $\calS_{\mathrm{init}} \subset \calS$, we first prompt an LLM to generate hypotheses for $\calS_{\mathrm{init}}$, which serve as our initial hypothesis bank $\mathcal{H}$. 
While initialized hypotheses may explain some portions of data, they often fall short of encompassing the full scope of the examples.
We thus introduce an update stage which serves a dual purpose: 
1) it increases the percentage of data explainable by the hypotheses and 2) it replaces any hypotheses that are found to be inaccurate.

In the update stage, for a training example 
$s$, we select the top $k$ high-reward hypotheses from the hypothesis bank $\mathcal{H}$.
The LLM is prompted to make a prediction with each of the top $k$ high-reward hypotheses on $s$.
Then we compute the accuracy of the inference and accordingly update the reward for each of the hypotheses.
If $w_{hyp}$ hypotheses predict incorrectly for the example $s$,
then $s$ is added to a wrong example pool $\mathcal{W}$.
Once the wrong example pool reaches a max size of $w_{max}$, the wrong examples in $\mathcal{W}$ are used to generate new hypotheses.
The wrong example pool represents the gap in knowledge that the current pool of hypotheses has for the dataset. Thus, by generating new hypotheses, the algorithm fills in these gaps. 
We update $\mathcal{H}$ with the newly generated hypotheses as per the rewards.

\para{Reward.}
As mentioned above, each hypothesis has an associated reward. 
In our algorithm, we 
use the reward function in the UCB algorithm due to similarities between the multi-arm bandit problem and our problem formulation. 
In particular, we consider each hypothesis to be an arm and each training example to be a ``pull''. We note, however, that unlike the multi-arm bandit problem, multiple hypotheses are tested for a singular train example.
Moreover, 
there can be new arms after hypotheses are updated, altering the setting from the standard static arms scenario to a dynamic arms scenario.
Formally, the reward is defined as
\begin{equation}
    r_i = \frac{\sum_{(x_j, y_j) \in \calS_i} I(y_j = \hat{y}_j)}{|\calS_i|} + \alpha\sqrt{\frac{\log{t}}{|\calS_i|}},
\end{equation}
where $\calS_i$ is the set of examples that have been used to evaluate the hypothesis $h_i$, $t$ is train time step, and $\alpha$ is a hyperparameter that controls the exploration term.
The first term in the reward function denotes the accuracy of the hypothesis for all $\calS_i$. 
The second term is the exploration term, which is computed based on the number of times the hypothesis has been selected and the number of training examples visited so far.
The accuracy term urges the algorithm to use well-performing hypotheses, whereas the exploration term encourages the algorithm to explore hypotheses that have not been selected many times.
Thus, the reward function strikes a balance between exploration and exploitation.

For more details on implementation of \ours, refer to \cref{appendix:hypogenic_details}.

\begin{algorithm}[t]
\caption{
    \ours
}
\label{alg:hyp_gen}
\small
\begin{algorithmic}[1]
\INPUT Training samples $\mathcal{S}, \mathtt{num\_init}, k, w_{max}, H$
\State // \textit{Initialize hypothesis bank}
\State $\mathcal{H}$ $\gets$ \textbf{generate\_hypotheses}($\{\mathcal{S}_i : i \leq \mathtt{num\_init} \}$) 
\State $\mathcal{W}$ $\gets$ \{\}
\For{$(x_t, y_t)\in \mathcal{S}$}
    \State $\mathcal{H}_\mathrm{top} \gets \{h : \text{$h \in \mathcal{H}$ has top k reward}\}$
    \For{$h \in \mathcal{H}_\mathrm{top}$} 
        \State $\hat{y}_t^h \gets$ \textbf{inference}($h$, $t$) 
        \State \textbf{update\_reward}($h, y_t, \hat{y}_t^h$)
    \EndFor
    \If {$|\{\mathbf{wrong}(\hat{y}_t^h): h \in \mathcal{H}\}| \geq w_{hyp}$} 
        \State // \textit{$w_{hyp}$ is dynamically determined, see \cref{appendix:hypogenic_details}}
        \State $\mathcal{W}$ $\gets$ $\mathcal{W}$ $\cup \{(x_t, y_t)\}$
    \EndIf

    \If {$|\mathcal{W}| = w_{max}$}
        \State $\mathcal{N} \gets$ \textbf{generate\_hypotheses}($\mathcal{W}$)
        \State $\mathcal{W}$ $\gets$ \{\}
        \State $\mathcal{H} \gets \{h : \text{$h \in \mathcal{H} \cup \mathcal{N}$ has top k reward}\}$
    \EndIf
\EndFor
\State \textbf{return} $\mathcal{H}$
\end{algorithmic}
\end{algorithm}

\subsection{Hypothesis-based Inference}
\label{sec:hypothesis-based-inference}

For efficiency purposes, we use each hypothesis on its own without accounting for their combinatorial effect during training; however, we should leverage the set of hypotheses as a whole during inference for at least two reasons. 
Firstly, some hypotheses may only apply to a subset of examples. Second, competing theories may require head-to-head comparisons.
Hence, we develop multiple inference strategies to account for these different styles of reasoning (see \cref{appendix:prompts} for prompts and \cref{appendix:inference_details} for implementation details).
\begin{itemize}%
    \item {\bf Best-accuracy hypothesis.} 
    The hypothesis $h$ with the highest accuracy from the hypothesis bank is included in the prompt to guide the model to perform inference. 
    
    \item {\bf Filter and weighted vote.}
    One hypothesis may not be enough to explain the data.
    Thus, this approach uses a combination of relevant hypotheses to make predictions for a single example.
    We first {\it filter} hypotheses by prompting an LLM to judge which hypotheses are relevant to the example.
    Next, an LLM is prompted to generate predictions for each of the relevant hypotheses, and these predictions are aggregated with {\it weighted vote}, where the weight is the training accuracy of the corresponding hypothesis. 
    
    \item {\bf Single-step adaptive inference.}
    Similar to {\it filter and weighted vote}, this approach leverages contextual information to choose hypotheses. The difference, however, is that it selects the most applicable hypothesis for each test example.
    Specifically, for a given test example, the LLM is tasked with identifying the most applicable hypothesis from a set of options. 
    For each hypothesis, we provide instances from the training set where the hypothesis was accurate. Then, the LLM selects the most relevant hypothesis by comparing the test example to these training examples and evaluating their similarity.
    Thereafter, we apply the hypothesis to the test example to perform inference. Please note that this is all done in one step with a long prompt.
    
    \item {\bf Two-step adaptive inference.}
    We divide the previous inference strategy into two steps:
    \begin{enumerate}[leftmargin=*,itemsep=-2pt,topsep=-2pt]
        \item The LLM determines the most relevant set of examples by comparing the test example with the corresponding examples of the hypotheses.
        \item Then, the corresponding hypothesis is provided to the LLM, which it uses to perform inference on the test example in a second prompt.
    \end{enumerate}
\end{itemize}

\section{Experiment Setup}
\label{sec:experiments}

We introduce the experiment setup to evaluate \ours.

\subsection{Tasks and Datasets}
\label{sec:datasets}

The choice of appropriate tasks is critical for evaluating the ability of LLMs to generate hypothesis.
The focus of our work is on generating hypotheses based on observed data.
A prerequisite is that potential hypotheses do exist.
In the context of classification, it implies that the classification performance is non-trivial. 
In addition, we need to ensure that the hypotheses describing the data are likely not a priori known by LLMs, which rules out standard tasks such as sentiment analysis.
Therefore, we use four datasets that satisfy these requirements: a synthetic task with a known true hypothesis and three \textit{real-world} datasets that exhibit complex underlying patterns and constitute widely studied social science problems.

{\bf \shoe} is a synthetic task we created to investigate the scenario where there is only one single valid hypothesis.
The task is to predict the color of the shoe that the customer will buy based on their appearance.
The input provides appearance features, namely, age, height, gender, color of the hat, color of the shirt, color of the bag, and size of the bag.
We construct this dataset such that the color of the shoe must match the color of the shirt.
Since there are six colors in total, this becomes a 6-class classification problem.

{\bf Deceptive review detection} is an instance of deception detection, a widely studied phenomenon in psychology and other social sciences~\citep{granhag2005deception}.
This particular task (\deceptive) requires distinguishing genuine reviews from fictitious ones \citep{ott2011finding}, where human performance is about chance~\citep{lai2019human}. 
The dataset includes 800 genuine reviews and 800 fictitious reviews for 20 hotels in Chicago.

{\bf Predicting popularity} is a notoriously challenging task in social sciences because it is known to be affected by seemingly random factors~\citep{salganik2006experimental}. 
We use two datasets in this work: \headline and \retweet.
\headline is derived from a dataset in the Upworthy Research Archive \citep{Matias2021TheUR}.
The original dataset was collected through A/B testing, where each user was shown pairs of a headline and image for multiple packages (articles).
Each user was exposed to only one of these pairs per package, and the clicks were recorded for each pair per package.\footnote{The Upworthy Research Archive only provides the image IDs instead of the graphics. We thus only use the headlines for our dataset.}
This process resulted in a total of 150,816 headlines across 22,666 packages. 
We construct a binary classification dataset by choosing the headlines that received the most clicks and least clicks for each package.
We remove all sets of duplicate headlines, which results in our version of the \headline dataset. 
The task for this dataset is to deduce which headline had more clicks in a pair. 
\retweet uses a dataset of 13,174 tweet pairs \citep{chenhao2014retweet}, which are matched by the topic and the author.
Similar to \headline, the task is to predict which one received more retweets.

\subsection{Baselines, Oracles, and Evaluation Metrics}
\label{sec:exp_setup}

We use three different LLMs in our experiments (\mixtral \citep{mixtral2023}, \gpt \citep{openai2023chatgpt}, and \claude \citep{anthropic2023claude2}).
We compare our approach with the following methods. 
\begin{enumerate}%

\item{\bf Zero-shot and few-shot prompting.} 
We provide LLMs with task-specific instructions (zero-shot), optionally accompanied by three demonstration examples (few-shot).

\item{\bf No updates.}
To assess the value of the update stage in our algorithm, we evaluate the performance of the initialized hypotheses. 
In particular, we pick the best-performing hypothesis on the training set and use it for inference on the test set.

\item{\bf Supervised Learning.}
We fine-tune \roberta \citep{Liu2019RoBERTaAR} and \llama \citep{Touvron2023Llama2O} on each of the datasets to serve as a non-interpretable oracle.
We include results for training on 200 examples and 1000 examples.
Since fine-tuning update model weights, we expect \roberta and \llama to set the upper bound on in-distribution datasets.

\end{enumerate}

We randomly sample 200 training examples and 300 test examples for each dataset.
Since all our datasets are classification tasks with ground truth labels, we use accuracy as our evaluation metric.
To understand the effect of the number of training examples, we evaluate the performance of all methods at 10, 25, 50, 100, and 200 training examples.
We also experiment with two different hypothesis bank sizes: 3 and 20 hypotheses to evaluate the impact of utilizing a larger number of hypotheses.
The detailed hyperparameters of our approach can be found in \cref{appendix:hyperparams}.

\section{Results}
\label{sec:results}

To demonstrate the effectiveness of our hypothesis generation approach,
we present results via three evaluation methods.
First, we show that in the standard supervised learning setup, our generated hypotheses enable more accurate predictions than baselines and even oracles when using a small set of examples.
Second, we evaluate the generated hypotheses by checking whether they can generalize across different inference LLMs and to out-of-distribution datasets.
We find surprisingly consistent performance even when using a different LLM to make inference from the generated hypotheses. 
So, we conduct a qualitative analysis to show that the generated hypotheses not only corroborate existing theories but also provide novel insights about the tasks at hand.  

\begin{table*}[t]
    \centering
    \resizebox{0.92\textwidth}{!}{%
    \begin{tabular}{@{}llrrrr@{}}
        \toprule
                           &                    & \textsc{Shoe}  & \textsc{Deceptive} & \textsc{Headline}   & \textsc{Tweet}        \\ 
        Models             & Methods            & \textsc{Sales} & \textsc{Reviews}   & \textsc{Popularity} & \textsc{Popularity}   \\ \midrule \midrule
        \roberta (Oracle) & Train 200            & 100.0      & 84.0              & 49.0       & 50.7      \\
                         & Train 1000           & 100.0      & 91.0              & 60.0       & 62.0      \\ \midrule
        \llama (Oracle)      & Train 200            & 100.0      & 88.7              & 49.7     & 50.3  \\ 
                          & Train 1000   & 100.0      & 92.3              & 60.0     & 51.3   \\ \midrule

		\claude			 & Zero shot			 & 36.0		 & 31.0				 & 59.0		 & 50.3		 \\ 
						 & Few shot				 & 75.0		 & 51.0				 & 60.0		 & 0.3*		 \\ 
						 & \ours (no updates)	 & 100.0		 & 70.3				 & 57.3		 & 59.0		 \\ 
						 & \ours			 & {\bf 100.0}		 & {\bf 75.3}				 & {\bf 61.3}		 & {\bf 62.0}		 \\ \midrule 
		\mixtral		 & Zero shot			 & 43.0		 & 55.0				 & 55.0		 & 2.7*		 \\ 
						 & Few shot				 & 79.0		 & 56.3				 & 55.3		 & 48.7		 \\ 
						 & \ours (no updates)	 & 96.0		 & 60.3				 & 59.7		 & 60.7		 \\ 
						 & \ours			 & {\bf 98.0}		 & {\bf 68.0}				 & {\bf 60.3}		 & {\bf 62.7}		 \\ \midrule 
		\gpt			 & Zero shot			 & 39.0		 & 50.0				 & 56.0		 & 41.0		 \\ 
						 & Few shot				 & 49.0		 & 55.0				 & 60.0		 & {\bf 62.0}		 \\ 
						 & \ours (no updates)	 & 100.0		 & 56.0				 & 44.0		 & 45.0		 \\ 
						 & \ours			 & {\bf 100.0}		 & {\bf 60.7}				 & {\bf 63.7}		 & 61.0		 \\ \bottomrule
    \end{tabular}
    }
    \caption{
        Prediction accuracies with 200 examples.
        We report the best numbers across all hyperparameter configurations, number of training examples, and inference strategies for \ours (we discuss their effect in details in \cref{sec:prediction_performance}).
        The sensitive nature of the \retweet dataset may cause models to have their safety mode triggered. These results are marked by * in the table.
    }
    \label{tab:model_performance}
\end{table*}

\subsection{Performance on Heldout Test Sets}
\label{sec:prediction_performance}
As discussed in the introduction, a side product of our approach is an interpretable hypothesis-based classifier.
We compare its performance with standard supervised learning with \finetune and few-shot in-context learning (\cref{tab:model_performance}).

\paragraph{Our generated hypotheses improve inference over standard zero-shot and few-shot inference.}
Across all LLMs, \ours outperforms the zero-shot learning by an average of 60\% on \shoe, 22.7\% on \deceptive, 5.1\% on \headline, and 30.6\% on \retweet.
Similarly, we find that \ours shows an increase from few-shot learning by 31.7\% on \shoe, 13.9\% on \deceptive, 3.3\% on \headline, and 24.9\% on \retweet.
Note that these results are inflated on \retweet as safety mode is triggered for \mixtral and \claude for zero-shot and few-shot learning respectively.
After computing the 95\% confidence intervals (with a binomial distribution assumption) for our results, the following results are significant for the real life datasets: \ours for \deceptive and \retweet with \claude and \mixtral, when comparing to their respective few shot baselines. 
If we relax the confidence interval to 90\%, the result for \headline with Mixtral is also statistically significant.
These results demonstrate that hypothesis-based inference can increase the performance of LLMs significantly. Further results can be found in \cref{tab:performance_with_CI}.
One exception is that our method performs slightly worse (by 1\%) than the few-shot baseline in the \retweet with \gpt.
One possible reason is that the few-shot demonstrations are effective at eliciting the pretraining knowledge in \gpt, possibly due to a large amount of tweets in pretraining data.
More detailed results are in \cref{appendix:detailed-results}. 

We also evaluate generated hypotheses with oracle inference, where the model retrospectively picks the best hypothesis for each prediction from the bank.
With oracle inference, \ours achieves on average 88.6\% on \deceptive, 84.1\% on \headline, and 88\% on \retweet across all LLMs, which are superior to results in \cref{tab:model_performance}.
This result further suggests that hypotheses generated by \ours are of high quality and can lead to accurate predictions when the correct hypothesis is selected.

\paragraph{\ours matches or even exceeds \finetune with the same number of training examples on most datasets.}
Both \ours and \finetune yield 100\% on the syntheic dataset. Moreover, \ours is 12.8\% and 11.2\% better than \roberta, and 12.1\% and 11.6\% better than \llama, on \headline and \retweet respectively with 200 training examples.
Since \finetune learns by updating model weights to minimize the cross-entropy loss, it tends to benefit from more training examples, so we increase training examples to 1000 for \finetune.
Despite the accuracy boost from more training examples, we find that \ours's best result still outperforms \roberta by 3.7\% and 0.7\%, and \llama by 3.7\% and 11.4\%, on \headline and \retweet, respectively.
One exception, however, is the \deceptive dataset.
We suspect that as word-level features are very useful in this dataset~\citep{ott2011finding},
they could be tougher for LLMs to extract but easier for fine-tuned models to grasp.

\paragraph{Updating the hypothesis bank leads to hypotheses of higher quality.} 
Comparing \ours with the ``no updates'' results, we find that updating hypotheses generally leads to better hypotheses, suggesting that our algorithm is effective at improving hypothesis quality.
The improvement is on average 0.7\% on \shoe, 5.8\% on \deceptive, 8.1\% on \headline, and 7\% on \retweet.
Another advantage of \ours over ``no updates'' is that sometimes the training examples exceed the context window size of LLMs, which can lead to degraded performance (\cref{fig:gpt_line_chart,fig:claude_line_chart}).

\begin{figure}[t]
    \centering
    \resizebox{0.95\linewidth}{!}{
    \includegraphics[width=0.8\textwidth]{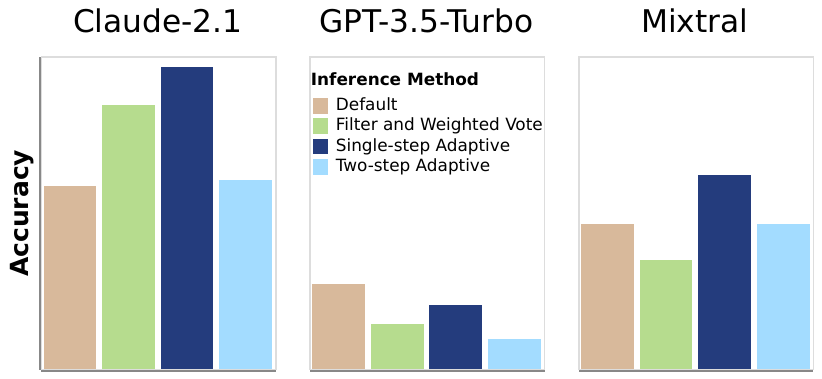}
    }
    \caption{
        \ours results with different inference strategies on \deceptive. 
        Single-step adaptive hypothesis-based inference is generally the most effective on this dataset.
    }
    \label{fig:hotel_inf_strategies}
\end{figure}

\paragraph{Effect of inference strategy.}
\cref{fig:hotel_inf_strategies} shows \ours results with different inference strategies on \deceptive.
Single-step adaptive inference is the most effective.
Generally, we find hypotheses to be one-sided, focusing on either characteristics of truthful or deceptive reviews. 
We thus need to consider more than one hypothesis to make a correct prediction, so best-accuracy hypothesis or two-step adaptive inference are not ideal.
On the other datasets, we find that the effect of inference strategy is much smaller (\cref{fig:inference_strategies}). 
Best-accuracy hypothesis is sufficient for \shoe and \headline, and filter and weighted vote works best for \retweet.
{\bf Whichever inference strategy we use, the trend of \ours against few-shot learning and \finetune remains largely the same.}

\paragraph{Generally, having more training examples and a larger hypothesis pool improves performance.} 
We show performance for different methods as number of training examples increase in \cref{fig:gpt_line_chart,fig:claude_line_chart,fig:mixtral_line_chart}.
We find \ours accuracy steadily increases as training size increases on \shoe, suggesting that an LLM is more likely to generate the best hypothesis given more examples.
For the real-world datasets, however, the performance sometimes peaks at training size at 25 or 100 before reaching to 200. 
We suspect that the evaluation of the hypothesis bank would be less stable for the real-world datasets, since more than one correct hypotheses are needed for the task.
We also find that using a hypothesis pool of size 20 leads to better performance than using a pool of size 3.

\textbf{Although this classification experiment is convenient to run and demonstrates that our generated hypotheses are reasonable, our main goal is to generate high-quality hypotheses rather than maximizing the performance of this particular way of using the hypotheses. The next two experiments are essential in understanding the quality of hypotheses through generalization and manual analysis.}

\subsection{Generalization of the Generated Hypotheses}

Our primary interest lies in the quality of the hypotheses. 
A good hypothesis should enable accurate inference by any AI model or even human and also generalize to unseen out-of-distribution dataset.
In this subsection, we mix and match different LLMs for generation and inference. 
We also evaluate the hypotheses in deceptive review prediction on a new out-of-distribution (OOD) dataset \citep{oodhotelreviews}.

\begin{table*}[t]
    \centering
    \resizebox{0.85     \textwidth}{!}{%
    \begin{tabular}{@{}llcccc@{}}
        \toprule

              &       & \textsc{Shoe}  & \textsc{Deceptive} & \textsc{Headline}   & \textsc{Tweet}        \\ 
        Generation Model           & Inference Methods        & \textsc{Sales} & \textsc{Reviews}   & \textsc{Popularity} & \textsc{Popularity}   \\ \midrule \midrule

        \claude & \claude & 100.0 & 67.3 & 57.7 & 62.0 \\
                & \mixtral & 94.0 & 65.0 & 57.7 & 59.3 \\
                & \gpt & 100.0 & 60.7 & 56.3 & 57.7 \\
        \midrule
        \mixtral & \claude & 99.0 & 69.7 & 59.0 & 58.7 \\
                 & \mixtral & 98.0 & 61.3 & 57.7 & 59.3 \\
                 & \gpt & 90.0 & 56.7 & 55.3 & 53.0 \\
        \midrule
        \gpt & \claude & 100.0 & 75.3 & 60.3 & 59.0 \\
             & \mixtral & 98.0 & 62.0 & 60.0 & 62.3 \\
             & \gpt & 100.0 & 57.3 & 58.7 & 56.3 \\
        \bottomrule

    \end{tabular}
    }
    \caption{
        Performance of cross-model generation and inference with train size = 200 using best-accuracy hypothesis inference and the best hypothesis bank size between 3 and 20.
    }
    \label{tab:cross_model}
\end{table*}

\paragraph{We find that the hypotheses generated by \ours generalize across models (\cref{tab:cross_model}).} 
Generally, we find \claude and \mixtral to be better at inference.
Thus, substituting the inference model with them lead to better performance for hypothesis generated with \gpt.
Substituting \claude and \mixtral as each other's inference model lead to small changes in performance.
On \shoe, the performance remains high for any inference model used.
\begin{table}[t]
    \centering
    \small
    \begin{tabular}{@{}llcccc@{}}
        \toprule
        Models           	 & \textsc{OOD} \\ \midrule \midrule
        RoBERTa (Oracle)	 & 73.0 (\decrease $11.0$) \\ 
        Llama-2-7B (Oracle)  & 78.7 (\decrease $10.0$) \\ \midrule
        \claude\ Few shot	 & 41.7 (\decrease $9.3$) \\ 
        \claude\ \ours		 & {\bf 74.7} (\increase $4.7$) \\ \midrule 
        \mixtral\ Few shot	 & 49.0 (\decrease $7.3$) \\ 
        \mixtral\ \ours		 & 64.7 (\increase $1.7$) \\ \midrule 
        \gpt\ Few shot		 & 52.0 (\decrease $3.0$) \\ 
        \gpt\ \ours			 & 60.7 (\increase $3.4$) \\ \bottomrule
        \end{tabular}
        \caption{
            Performance on OOD deceptive reviews.
        }
        \label{tab:ood_results}
\end{table}

Performance even increases for \deceptive and \headline when using \claude as the inference model.
For the cases where performance drops from \claude to \mixtral, the decrease is marginal: 2.3\% on \deceptive and 2.7\% on \retweet.

These results suggest that the hypotheses generated by \ours are generalizable across different LLMs, which somewhat contradicts the claim in \citet{qiu2024phenomenal} that LLMs cannot reliably interpret the hypotheses.
We suspect that the reason is that our tasks only rely on natural language, while their tasks rely on notions of worlds and are fed into symbolic interpreters.

\paragraph{Our generated hypotheses generalize to an out-of-distribution dataset.}
\cref{tab:ood_results} presents an overview for the OOD deceptive review dataset. 
This dataset differs from \deceptive by including reviews from 
four cities sourced from different websites~\citep{oodhotelreviews}.
We find that \ours outperforms few-shot learning by an average of 19.1\%. 
Despite the distribution shift, \ours
surprisingly increases accuracy from \deceptive by an average of 3.3\%, suggesting our hypotheses generalize well to this OOD dataset.
\claude remains the best performing model.
In comparison, 
the performance of \roberta drops by 11\%, and \llama drops by 10\%.
As a result, \ours with \claude outperforms \roberta by 1.7\%, demonstrating the robustness of hypothesis-based inference.
Refer to \cref{sec:full_ood_results} for more details.

\subsection{Qualitative Analysis}
\label{sec:analysis}

For the synthetic dataset, all models are able to find the true underlying hypothesis for \shoe: ``customers tend to buy shoes that match the color of their shirt.''
For the real-world datasets, we search for studies on these datasets on Google Scholar and compare our hypotheses with findings from the literature. We confirm the validity of some of our hypotheses and discover new insights about the tasks that previous studies did not touch upon.
We show a few examples in \cref{tab:hypotheses_analysis_small}, and the full list of hypotheses can be found in \cref{appendix:hypothesis-analysis}.

\paragraph{Our generated hypotheses align with useful features in existing literature.} 
For \deceptive, we find that deceptive reviews are more likely to be emotional, use superlatives, or contain information that could not have been directly experienced. Similar findings are also found by previous studies on \deceptive \citep{lai2020chicago,anderson2014reviews,ott2011finding,li2014generalrule}. 
For \retweet, we discover that tweets that are concise, with specific or relevant hashtags, or with emotional tones are more likely to be retweeted more, aligning with prior studies~\citep{chenhao2014retweet,gligoric2019causal}.
For \headline, we find that 
revealing something new or using vivid language and imagery can 
drive engagement from readers to click on headlines. Previous studies also find these rules apply to online news headlines \citep{banerjee2023language, sadoski2000engaging}.
\begin{table*}[t!]
\centering
\resizebox{0.95\textwidth}{!}{%
\begin{tabular}{@{}lp{8cm}p{4.3cm}@{}}
\toprule
\textbf{Dataset} & \textbf{Finding} & \textbf{Supported/Novel} \\ \midrule \midrule
\deceptive & Deceptive reviews contain more emotional terms. & \citet{li2014generalrule} \\ 
           & Truthful reviews would mention weddings or special occasions. & \novel \\
\midrule
\headline  & Using vivid language and imagery helps. & \citet{banerjee2023language} \\
           & Headlines that frame the content in a personal or relatable way are clicked more. & \novel \\
\midrule
\retweet   & Tweets with emotional tones are retweeted more. & \citet{chenhao2014retweet} \\
           & Mentioning influential individuals or organizations leads to more retweets. & \novel \\
           
\bottomrule

\end{tabular}
}
\caption{Selected examples of generated hypotheses (on the real-world datasets) and whether they support existing findings or are novel.}
\label{tab:hypotheses_analysis_small}
\end{table*}

\paragraph{We also discover new insights with our generated hypotheses.}
For the \deceptive dataset, truthful reviews could mention the reviewer's purpose for staying at the hotel (e.g., business trip, vacation), but deceptive ones tend not to have this information. 
For \headline, we find that headlines that frame the content in a personal or relatable way are clicked more.
For \retweet, tweets that mention influential individuals or organizations are more likely to be retweeted.

\paragraph{Intriguingly, one of our hypotheses contradicts a feature engineering result.} 
\citet{ott2011finding} find that the token ``future'' is associated with deceptive reviews, while one of our hypotheses says that mentions of ``past experiences or future travel plans'' are indicative of truthfulness. This discrepancy is interesting, because the context for the token ``future'' is unclear. 
It could be in the context of future plans but could also be as a complaint about ``never going to stay at the hotel in the future.''
Feature engineering is limited by contextual ambiguity, whereas our generated hypotheses and their interpretation by LLMs overcome such limitations.

\paragraph{Our automatic evaluation of hypothesis quality also reflects negative findings.}
Given mixed evidence from previous literature on the effect of ``reading ease'' on headline clicks,
\citet{banerjee2023language} 
finds that reading ease negatively impacts click-through rates in \headline through careful feature engineering.
Consistent with this result, 
we found that the hypotheses that claim ``straightforward'' and ``clear'' writing to be indicative of higher click-through rates have relatively lower accuracies during training.

\section{Additional Related Work}
\label{sec:related-work}

\paragraph{Concept/pattern discovery.}
Our work is connected to many recent studies on using LLMs to propose ``hypotheses'', notably, \citet{qiu2024phenomenal} and \citet{zhong2023goal}.
\citet{qiu2024phenomenal} is motivated by testing the ability of LLMs to perform human-like induction reasoning,
and \citet{zhong2023goal} aims to support open-ended exploration. Similar to \citet{qiu2024phenomenal}, \citet{Tenenbaum2011HowTG} is motivated by human inductive reasoning and examines concept induction in synthetic settings. 
\citet{ellis2020dreamcoder} further 
learns to program concepts. \citet{yang2024inductivereasoners} performs LLM-based inductive reasoning with a dataset that requires existing fact-rule pairs, which is not applicable in our real-world problems.
\citet{funsearch} generates programs that lead to mathematical discovery.
Similar to \citet{zhong2023goal}, \citet{pham2023topicgpt} generates and refine a list of topics to achieve interpretable topic modeling for open-ended exploration.
\citet{honovich2022instruction} explores the deduction of task description from examples. 
Additionally, \citet{qi2023large}, \citet{wang2024scimon}, and \citet{baek2024researchagent} use LLMs to generate hypotheses from previous literature. \citet{yang2024moose} tries to generate hypotheses from raw web corpus, but their method is not automated or scalable as it requires human annotated hypotheses from existing literature.
Our work, in contrast, focuses on hypothesis generation between the input and the label for real-world challenging tasks and uses a UCB-style reward to propose novel algorithms.

\paragraph{Reasoning with LLMs.}
Although it is not our primary goal, our results show that hypothesis-based classifiers can outperform few-shot prompting.
As hypotheses may be viewed as a form of reasoning, it is related to reasoning with LLMs
\citep[][{\it i.a.}]{weiChainThoughtPrompting2022,wang2023selfconsistency}.
In particular, our work differs from chain-of-thought reasoning because 
no predefined reasoning structure is available.
Moreover, an important distinction between reasoning and hypothesis generation is that the former leverages established reasoning, while the latter requires both proposition and verification of the hypotheses, to discover unknown knowledge.

\paragraph{LLMs for (social) sciences.}
Increasing attention has been brought to the use of LLMs in social science research~\citep[][{\it i.a.}]{ziems2024can,kim2023aiaugmented}.
Our experiments demonstrate the potential of LLMs in generating hypotheses for social science research to discover unknown knowledge in the data.
Furthermore, our approach can be extended to natural sciences for general scientific discovery.

\section{Conclusion \& Further Discussion}

In this work, we propose \ours, a novel data-driven and automated method that leverages LLMs to generate hypotheses with the goal of discovering unknown knowledge. 
With \ours, we are not only are able to generate human-interpretable hypotheses but also achieve better predictive performance against competitive baselines and even oracles. Furthermore, our method can generalize well with different models and datasets, including open models. Notably, with our generated hypotheses, we uncover new insights in real-world tasks that are widely studied in social sciences. 

The key to success in \ours is not that LLMs remembers the correct hypotheses, but lies in their ability to ``hallucinate'' and combine potentially relevant concepts.
The exploration-exploitation process then identifies the valuable hypotheses.
\ours can be directly applied to complex social science tasks.
We encourage future work to explore hypothesis generation that requires additional modalities and/or leverages existing literature along with past observations.

\section{Limitations}
\label{sec:limitations}

We address common concerns using a Q\&A format.

\textbf{Q:} Why only experiment with social science tasks?

\textbf{A:} Math and physics problems and hypotheses are hard to represent in natural language and usually require symbolic parsers \citep{alphageom}. 
We leverage LLMs to perform tasks that it is naturally adept at, which lead us to social science tasks. We find that \ours demonstrates strong results for the selected tasks, indicating new possibilities in using LLMs for scientific discovery. We leave extending our framework to natural science tasks as future work.

\textbf{Q:} Why is \ours effective, given that the accuracy improvement is not significant in some settings?

\textbf{A:} Even if there is no significant improvement in accuracy, the benefits of \ours are found in the quality of hypotheses. We find that the generated hypotheses discover new patterns that were previously unseen, as discussed in \cref{sec:analysis}. 
Additionally, it is worth noting that LLMs are imperfect at reasoning.
Thus, hypothesis-based inference with LLMs may not accurately reflect the quality of the hypotheses.

\textbf{Q:} Since you worked on some old datasets, what if the LLMs have pre-trained knowledge about these tasks?

\textbf{A:} In \cref{tab:model_performance}, the zero/few-shot learning results suggest that the models cannot solve the tasks by memorizing the data. Additionally in \cref{sec:analysis}, we show that \ours reveal new hypotheses, based on the literature space that we can manually search. Even if the models have been pre-trained on the datasets, these hypotheses were not reported in previous literature. This suggests that even experienced researchers still struggle in finding the hypotheses that \ours generate. 

\textbf{Q:} What hyperparameters have you tried?

\textbf{A:}
We aim to provide a robust framework for hypothesis generation, as opposed to focusing on the optimization of results. 
Thus, we did not perform an extensive hyperparameter search with the generation portion of \ours. 
We did not adjust the value of $k$, which determines $\mathcal{H}_{\mathrm{top}}$ in \cref{alg:hyp_gen} to maintain efficiency.
Additionally, we only considered the effect of using a hypothesis bank size of $3$ and $20$ to only test using an extremely small hypothesis bank size and a large one.
The ideal hypothesis bank size may require further investigation. 
Finally, we only tested the size of our wrong example bank $w_{max}$ as $10$ to strike a balance between context window sizes and generation of good quality hypotheses. 
We believe that a more thorough hyperparameter search could improve the performance of our methodology.

\textbf{Q:} How costly is your approach?

\textbf{A:} \ours has high latency, specifically when using inference methods that require multiple prompts. 
For example, the filter and weighted vote inference policy requires iterating through the top hypotheses to determine relevance and then performing inference if it is relevant.
For single-step adaptive inference and best accuracy hypothesis, however, \ours is efficient. 
Given that we request reasoning for all inference prompts, the procedure can be time-consuming and require financial costs (e.g., \gpt takes \$2.05 on average over 76 experiments with an average of 1.5 hours per experiment).
This concern is alleviated when using open models.
However, all these processes are still relatively cheap compared to human efforts.

\textbf{Q:} What are some potential risks of hypothesis generation?

\textbf{A:} One potential risk of hypothesis generation is that there is little guard regarding steorotypes and biases being confirmed if given data that may seem to enforce them.
As a result, it can be potentially harmful to use \ours in a real-world setting without proper oversight.
Additionally, if the data reveals personal information regarding people, there is no guarantee that the hypotheses generated will not reveal this information.
We highly recommend human-AI collaboration in using \ours to ensure that the generated hypotheses are ethical and unbiased.

\section*{Acknowledgments}
We thank the anonymous reviewers for their suggestions.
We also thank members of the Chicago Human+AI Lab for their helpful comments.
This work is supported in part by NSF grants IIS-2126602 and an award from the Sloan Foundation.

\bibliography{custom.bib}

\clearpage
\appendix

\section{Prompts}
\label{appendix:prompts}
We follow the general prompt engineering guide from Claude \citep{anthropic2023claude2} to craft the prompts.
Specifically for all the prompts we use for LLMs, we split them into instruction and user prompts. In the instruction prompt, we first set a tone and context, followed by an explicit task description, and then specify the answer format. The user prompt then includes useful information such as past examples and learned hypothesis. By the end of the user prompt, we ask the LLM to make a prediction. At generation time, we input the instruction prompt to LLMs as system prompt, wrapped by the corresponding system prompt tokens for each model. 
Below are some example templates for the prompts associated with each task.
\subsection{Shoe Sales}
\label{appendix:prompts:shoe}

\begin{lstlisting}[caption={Hypothesis Generation.},label={lst:hypgen:shoe},firstnumber=auto]
@\textcolor{mygray}{Instruction Prompt}@
You're a helpful assistant. Your task is given as follows:
Given a set of observations, we want to generate hypotheses that are useful for predicting the color of the shoes given the appearance of the person. 
Please be concise and keep the hypotheses to be one-sentence long.  
Please generate them in the format of 
{1. [hypothesis]. 
2. [hypothesis].
... 
@\textcolor{mypurple}{<num\_hypotheses>}@. [hypothesis].}
Only propose @\textcolor{mypurple}{<num\_hypotheses>}@ possible hypotheses in total.
No need to explain the hypotheses.

@\textcolor{mygray}{User Prompt}@
We made some observations:
@\textcolor{mygray}{··· more examples here ···}@
Based on the above observations, generate @\textcolor{mypurple}{<num\_hypotheses>}@ hypotheses.
Please be concise and keep the hypotheses to be one-sentence long.  
Please generate them in the format of 
{1. [hypothesis]. 
2. [hypothesis].
... 
@\textcolor{mypurple}{<num\_hypotheses>}@. [hypothesis].}
Only propose @\textcolor{mypurple}{<num\_hypotheses>}@ possible hypotheses in total.
\end{lstlisting}

\begin{lstlisting}[caption={Hypothesis-based Inference.},label={lst:inference:shoe},firstnumber=auto]
@\textcolor{mygray}{Instruction Prompt}@
You are a shoe salesman and want to recommend shoes to customers. There are white, red, orange, green, blue, and black shoes.
From past experiences, you learned some patterns. 
Now, at each time, you should apply the learned pattern, given below, to a new customer and recommend a shoe color.
Give an answer for the shoe color recommendation. The answer should be one color word. It has to be one of white, red, orange, green, blue, and black.

@\textcolor{mygray}{User Prompt}@
Our learned pattern: @\textcolor{mypurple}{<hypothesis\_high\_reward>}@
New customer: @\textcolor{mypurple}{<appearance>}@ is buying a pair of shoes, the shoes should be which color?
Answer: 
\end{lstlisting}

\begin{lstlisting}[caption={Zero/\textcolor{myorange}{Few}-shot Inference.},label={lst:fewshot:shoe},firstnumber=auto]
@\textcolor{mygray}{Instruction Prompt}@
You are a shoe salesman and want to recommend shoes to customers. There are white, red, orange, green, blue, and black shoes. 
Give your answer for the shoe color recommendation. The answer should be one color word. It has to be one of white, red, orange, green, blue, and black. If you do not have enough information to make a recommendation, you should give the answer "unknown". 
Give your final answer in the format of "Final answer: [answer]."

@\textcolor{mygray}{User Prompt}@
@\textcolor{myorange}{Here are some examples of customers with certain features buying certain products:}@
@\textcolor{myorange}{··· more examples here ···}@
New customer: @\textcolor{mypurple}{<appearance>}@ is buying a pair of shoes, the shoes should be which color?
Answer:
\end{lstlisting}

\begin{lstlisting}[caption={Example-based Hypothesis Selection and Inference. \textcolor{mypurple}{<adaptive\_info\_prompt>} consists of several hypotheses and the corresponding examples they got correct during generation time.},label={lst:adaptive:shoe},firstnumber=auto]
@\textcolor{mygray}{Instruction Prompt}@
You are a shoe salesman and want to recommend shoes to customers. There are white, red, orange, green, blue, and black shoes.
From past experiences, you learned some patterns. 
For each pattern, you will also see a couple of examples that worked for each pattern.
Choose a pattern. To do this, look at the examples of each pattern, and see which of the examples the current customer is closest to. Choose the pattern corresponding to that example.
Give an answer for the shoe color recommendation. 
The answer should be one word. It has to be one of white, red, orange, green, blue, and black.
Give your final answer in the following format:
Reasoning for choosing pattern: reason,
Chosen pattern: pattern,
Reasoning for choice of prediction: reason,
Final Answer: answer

@\textcolor{mygray}{User Prompt}@
Here are some previously generated patterns with some example where it predicted correcly what color of shoe the customer bought.
@\textcolor{mypurple}{<adaptive\_info\_prompt>}@
New customer: @\textcolor{mypurple}{<appearance>}@ is buying a pair of shoes, the shoes should be which color?
Answer: 
\end{lstlisting}

\subsection{Deceptive Reviews}
\label{appendix:prompts:hotel_reviews}
\begin{lstlisting}[caption={Hypothesis Generation.},label={lst:hypgen:hotel},firstnumber=auto]
@\textcolor{mygray}{Instruction Prompt}@
You're a professional hotel review analyst.
Given a set of hotel reviews, we want to generate hypotheses that are useful for predicting whether a review is truthful or deceptive. In other words, we want to know whether the review is written by a someone who actually lived in the hotel.
Using the given examples, please propose @\textcolor{mypurple}{<num\_hypotheses>}@ possible hypothesis pairs.
These hypotheses should identify specific patterns that occur across the provided reviews.
Each hypothesis should contain a pair of the following:
1. A hypothesis about what makes reviews more likely to be truthful
2. The opposite hypothesis about what makes reviews more likely to be deceptive 
Generate them in the format of 1. [hypothesis], 2. [hypothesis], ... @\textcolor{mypurple}{<num\_hypotheses>}@. [hypothesis].
The hypotheses should analyze what kind of reviews are likely to be truthful or deceptive.

@\textcolor{mygray}{User Prompt}@
We have seen some hotel reviews:
@\textcolor{mygray}{··· more examples here ···}@
Please generate hypotheses that are useful for predicting whether a review is truthful or deceptive. 
Propose @\textcolor{mypurple}{<num\_hypotheses>}@ possible hypotheses. Generate them in the format of 1. [hypothesis], 2. [hypothesis], ... @\textcolor{mypurple}{<num\_hypotheses>}@. [hypothesis].
Proposed hypotheses:
\end{lstlisting}

\begin{lstlisting}[caption={Hypothesis-based Inference.},label={lst:inference:hotel},firstnumber=auto]
@\textcolor{mygray}{Instruction Prompt}@
You are a professional deceptive detection agent and your job is to determine whether a hotel review is truthful or deceptive. 
In other words, we want to know whether the review is written by someone who had real experiences with the hotel. 
From past experiences, you learned a pattern. 
You need to determine whether each of the patterns holds for the current hotel review, and also predict whether the current hotel review is truthful or deceptive. 
Give an answer. The answer should be one word (truthful or deceptive).
Give your final answer in the format of {Final answer: answer}

@\textcolor{mygray}{User Prompt}@
Our learned pattern: @\textcolor{mypurple}{<hypothesis\_high\_reward>}@
A hotel review is the following: @\textcolor{mypurple}{<review>}@
Given the pattern you learned above, give an answer of whether the hotel review above is deceptive or truthful.
Think step by step.
First step: Think about which pattern can be applied to the hotel review.
Second step: Based on the pattern, is this hotel review deceptive or truthful?
\end{lstlisting}

\begin{lstlisting}[caption={Zero/\textcolor{myorange}{Few}-shot Inference.},label={lst:fewshot:hotel},firstnumber=auto]
@\textcolor{mygray}{Instruction Prompt}@
You are a deceptive detection agent and want to determine whether a hotel review is truthful or deceptive.
In other words, we want to know whether the review is written by a someone who actually lived in the hotel.
You need to determine whether this pattern holds for the current hotel review, and also predict whether the current hotel review is truthful or deceptive.
Give an answer. The answer should be one word (truthful or deceptive).

@\textcolor{mygray}{User Prompt}@
@\textcolor{myorange}{We have seen some hotel reviews:}@
@\textcolor{myorange}{··· more examples here ···}@
A hotel review is the following: @\textcolor{mypurple}{<review>}@
Is this hotel review truthful or deceptive?
Answer:
\end{lstlisting}

\begin{lstlisting}[caption={Example-based Hypothesis Selection and Inference. \textcolor{mypurple}{<adaptive\_info\_prompt>} consists of several hypotheses and the corresponding examples they got correct during generation time.},label={lst:adaptive:hotel},firstnumber=auto]
@\textcolor{mygray}{Instruction Prompt}@
You are a professional hotel review analyst and you are able to determine whether a hotel review is deceptive or truthful. 
In other words, your job is to analyze if a hotel review review is written by someone who had genuine experiences with the hotel. 
From past experiences, you learned some patterns. 
For each pattern, you will also see a couple of examples that worked for each pattern.
First step: take a careful look at the examples associated with each pattern, and see which set of examples the current hotel review is most similar with. Choose and repeat the pattern corresponding to that examples set.
Next, apply the pattern on the new sample to determine whether the new hotel review is deceptive or truthful.
Finally, give an answer. The answer should be one word (deceptive or truthful). 
Please give your final answer in the following format:
Reasoning for choosing pattern: reason,
Chosen pattern: pattern,
Reasoning for choice of prediction: reason,
Final Answer: answer

@\textcolor{mygray}{User Prompt}@
Here are some previously generated patterns with some example where it predicted correctly if a hotel review is deceptive or truthful.
@\textcolor{mypurple}{<adaptive\_info\_prompt>}@
A hotel review is the following: @\textcolor{mypurple}{<review>}@
Is this hotel review truthful or deceptive?
Think step-by-step.
Step 1: Look at the new hotel review and compare it with the set of examples associated with each provided pattern. 
Step 2: Find the set of examples that is the most similar to the new hotel review, pick and repeat the pattern associated with that set of examples.
Step 3: Apply the pattern you picked to the new hotel review and predict whether the new hotel review is deceptive or truthful.
Step 4: Give your final answer.
Answer:
\end{lstlisting}

\subsection{Headlines With More Clicks}
\label{appendix:prompts:headline_binary}
\begin{lstlisting}[caption={Hypothesis Generation.},label={lst:hypgen:headline},firstnumber=auto]
@\textcolor{mygray}{Instruction Prompt}@
You are a professional writer for an online newspaper company. 
Given a pair of headlines created for the same article, you are asked to determine which will get more clicks. It is likely that the pair of headlines shares similarities, so please focus on their differences. 
What difference in two headlines leads to more clicks on one than the other?
You will be given a set of observations of the format:
Headline 1: [headline]
Headline 2: [headline]
Observation: [observation].
Based on the observations, please generate hypotheses that are useful for explaining why one headline out of the pair gets more clicked than the other.
These hypotheses should identify patterns, phrases, wordings etc. that occur across the provided examples. They should also be generalizable to new instances.
Please propose @\textcolor{mypurple}{<num\_hypotheses>}@ possible hypotheses and generate them in the format of 1. [hypothesis], 2. [hypothesis], ... @\textcolor{mypurple}{<num\_hypotheses>}@. [hypothesis].

@\textcolor{mygray}{User Prompt}@
Here are the observations:
@\textcolor{mygray}{··· more examples here ···}@
Please generate hypotheses that can help determine which headlines have more clicks.
Please propose @\textcolor{mypurple}{<num\_hypotheses>}@ possible hypotheses.
Generate them in the format of 1. [hypothesis], 2. [hypothesis], ... @\textcolor{mypurple}{<num\_hypotheses>}@. [hypothesis]. 
Proposed hypotheses:
\end{lstlisting}

\begin{lstlisting}[caption={Hypothesis-based Inference.},label={lst:inference:headline},firstnumber=auto]
@\textcolor{mygray}{Instruction Prompt}@
You are a professional writer for an online newspaper company. 
Given a pair of headlines created for the same article, you are asked to determine which will get more clicks. It is likely that the pair of headlines shares similarities, so please focus on their differences. 
From past experiences, you learned some patterns.
Now, at each time, you should apply the learned pattern to a new pair of headlines that are created for a new article and determine which headline gets clicked more.
The answer for the higher clicks should be in the form "Headline _" where _ is either 1 or 2.
Please give your final answer in the format of {Final Answer: Headline _.}

@\textcolor{mygray}{User Prompt}@
Learned pattern: @\textcolor{mypurple}{<hypothesis\_high\_reward>}@
Given the pattern you learned above, predict which of the following headlines will get more clicks:
Headline 1: @\textcolor{mypurple}{<headline\_1>}@
Headline 2: @\textcolor{mypurple}{<headline\_2>}@
Think step by step.
Step 1: Think about whether the pattern can be applied to the headlines.
Step 2: Analyze the difference between "Headline 1" and "Headline 2".
Step 3: Based on the pattern, which headline is likely to get more clicks?
\end{lstlisting}

\begin{lstlisting}[caption={Zero/\textcolor{myorange}{Few}-shot Inference.},label={lst:fewshot:headline},firstnumber=auto]
@\textcolor{mygray}{Instruction Prompt}@
YYou are a writer for an online newspaper company. So you are excellent at determining which headlines are more likely to cause users to click on the article.
You will be given two headlines, and determine which headline was clicked more often.
You are only to give your answer.
The answer for the higher clicks should be of the form "Headline _" where _ is either 1 or 2. 
Give your final answer in the following format:
"Answer: Headline _"

@\textcolor{mygray}{User Prompt}@
@\textcolor{myorange}{Here are some previous examples to help you:}@
@\textcolor{myorange}{··· more examples here ···}@
Which of the following headlines has more clicks:
Headline 1: @\textcolor{mypurple}{<headline\_1>}@
Headline 2: @\textcolor{mypurple}{<headline\_2>}@
\end{lstlisting}

\begin{lstlisting}[caption={Example-based Hypothesis Selection and Inference. \textcolor{mypurple}{<adaptive\_info\_prompt>} consists of several hypotheses and the corresponding examples they got correct during generation time.},label={lst:adaptive:headline},firstnumber=auto]
@\textcolor{mygray}{Instruction Prompt}@
You are a professional writer for an online newspaper company.
You are excellent at determining which headlines are more likely to be clicked by users.
From past experiences, you learned some patterns.
For each pattern, you will also see a couple of examples that worked for each pattern.
Please choose a pattern. To do this, look at the examples associated with each pattern, and find which set of the examples are closest to the given pair of headlines. 
Please choose the pattern corresponding to that set of examples.
The answer for the higher clicks should be of the form "Headline _" where _ is either 1 or 2. 
Please give your final answer in the following format:
Reasoning for choosing pattern: reason,
Chosen pattern: pattern,
Reasoning for choice of prediction: reason,
Final Answer: answer

@\textcolor{mygray}{User Prompt}@
Here are some previously generated patterns with some examples where it predicted which one of the pair of headlines got more clicks.
@\textcolor{mypurple}{<adaptive\_info\_prompt>}@
Which one out of the following pair of headlines will get more clicks?
Headline 1: @\textcolor{mypurple}{<headline\_1>}@
Headline 2: @\textcolor{mypurple}{<headline\_2>}@
Think step by step.
Step 1: Look at the new pair of headlines and compare them with the examples associated with each pattern.
Step 2: Find the set of examples that is closest to the given pair of headlines, and pick the pattern associated with that set of examples.
Step 3: Apply the picked pattern to the new pair of headlines. Based on that pattern, think about which one out of the pair of headlines will get more clicks.
Step 4: Give your final answer.
\end{lstlisting}

\subsection{Retweeted More}
\label{appendix:prompts:retweet}
\begin{lstlisting}[caption={Hypothesis Generation.},label={lst:hypgen:retweet},firstnumber=auto]
@\textcolor{mygray}{Instruction Prompt}@
You are a social media expert. You are an expert at determining which tweet will be retweeted more.
Given a set of observations, you want to generation hypotheses that will help predict which tweet out of a pair of tweets is more likely to be retweeted.
Please note that the paired tweets are about the same content and are posted by the same user, so you should focus on the wording difference between the two tweets in each pair.
Please propose @\textcolor{mypurple}{<num\_hypotheses>}@ possible hypotheses.
Please generate them in the format of 1. [hypothesis], 2. [hypothesis], ... @\textcolor{mypurple}{<num\_hypotheses>}@. [hypothesis].
Please make the hypotheses general enough to be applicable to new observations.

@\textcolor{mygray}{User Prompt}@
We made some observations:
@\textcolor{mygray}{··· more examples here ···}@
Generate hypotheses that are useful for predicting which tweet out of a pair of tweets is more likely to be retweeted.
Please note that the paired tweets are about the same content and are posted by the same user, so you should focus on the wording difference between the two tweets in each pair.
Please propose @\textcolor{mypurple}{<num\_hypotheses>}@ possible hypotheses. 
Please generate them in the format of 1. [hypothesis], 2. [hypothesis], ... @\textcolor{mypurple}{<num\_hypotheses>}@. [hypothesis].
Proposed hypotheses:
\end{lstlisting}

\begin{lstlisting}[caption={Hypothesis-based Inference.},label={lst:inference:retweet},firstnumber=auto]
@\textcolor{mygray}{Instruction Prompt}@
You are a social media expert.
Given a pair of tweets, you are asked to predict which tweet will be retweeted more.
Please note that the paired tweets are about the same content and are posted by the same user, so you should focus on the wording difference between the two tweets.
From past experiences, you learned a pattern.
Now, at each time, you should apply a learned pattern to a pair of tweets and determine which one will get more retweets. 
The answer for the higher retweets should be of the form "the _ tweet" where _ is either first or second. 
Please give your final answer in the format of {Final answer: the _ tweet}

@\textcolor{mygray}{User Prompt}@
Our learned pattern: @\textcolor{mypurple}{<hypothesis\_high\_reward>}@
The first tweet: @\textcolor{mypurple}{<first\_tweet>}@
The second tweet: @\textcolor{mypurple}{<second\_tweet>}@
Given the pattern you learned above, predict which one of the two tweets will get more retweets.
Think step by step.
First step: Think about if the pattern can be applied to the tweets.
Second step: Analyze the textual difference between the two tweets.
Third step: Based on the pattern, which tweet is more likely to get more retweets?
Final step: Give your final answer in the format of {Final answer: the _ tweet}
Final answer:
\end{lstlisting}

\begin{lstlisting}[caption={Zero/\textcolor{myorange}{Few}-shot Inference.},label={lst:fewshot:retweet},firstnumber=auto]
@\textcolor{mygray}{Instruction Prompt}@
You are a social media expert.
Given a pair of tweets, you are asked to predict which tweet will be retweeted more.
Please note that the paired tweets are about the same content and are posted by the same user, so you should focus on the wording difference between the two tweets.
The answer for the higher retweets should be of the form "the _ tweet" where _ is either first or second. 
Please give your final answer in the format of {Final answer: the _ tweet}

@\textcolor{mygray}{User Prompt}@
@\textcolor{myorange}{Here are some examples:}@
@\textcolor{myorange}{··· more examples here ···}@
The first tweet: @\textcolor{mypurple}{<first\_tweet>}@
The second tweet: @\textcolor{mypurple}{<second\_tweet>}@
Which one of the two tweets will get more retweets?
\end{lstlisting}

\begin{lstlisting}[caption={Example-based Hypothesis Selection and Inference. \textcolor{mypurple}{<adaptive\_info\_prompt>} consists of several hypotheses and the corresponding examples they got correct during generation time.},label={lst:adaptive:retweet},firstnumber=auto]
@\textcolor{mygray}{Instruction Prompt}@
You are a social media expert.
Given a pair of tweets, you are asked to predict which tweet will be retweeted more.
Please note that the paired tweets are about the same content and are posted by the same user, so you should focus on the wording difference between the two tweets.
From past experiences, you learned some patterns.
You should apply a learned pattern to a pair of tweets and determine which one will get more retweets. 
For each pattern, you will also see a couple of examples that worked for each pattern.
Please choose a pattern. To do this, look at the examples associated with each pattern, and find which set of the examples are closest to the given pair of tweets. 
Please choose the pattern corresponding to that set of examples.
Please give your final answer in the following format:
Reasoning for choosing pattern: reason,
Chosen pattern: pattern,
Reasoning for choice of prediction: reason,
Final Answer: answer

@\textcolor{mygray}{User Prompt}@
Here are some previously generated patterns with some examples where it predicted which tweet will will be retweeted more.
@\textcolor{mypurple}{<adaptive\_info\_prompt>}@
The first tweet: @\textcolor{mypurple}{<first\_tweet>}@
The second tweet: @\textcolor{mypurple}{<second\_tweet>}@
Which one of the two tweets will get more retweets?
Think step by step.
Step 1: Look at the new pair of tweets and compare them with the examples associated with each pattern.
Step 2: Find the set of examples that is closest to the given pair of tweets, and pick the pattern associated with that set of examples.
Step 3: Analyze the textual difference between the two tweets.
Step 4: Apply the picked pattern to the new pair of tweets. Based on that pattern, think about which one out of the pair of headlines will get more clicks.
Step 5: Give your final answer.
\end{lstlisting}

\section{Implementation and Setup Details}
\label{appendix:implementation-details}

\subsection{\ours implementation}
\label{appendix:hypogenic_details}
\paragraph{Sampling} When initializing the rewards of newly generated hypotheses, we use the examples in the wrong example bank to do so. 
Given that we work in a low data regime, for hypotheses generated near the end of the training loop, the accuracies of hypotheses are likely to be biased.
To counter this phenomenon, we also allow for the hypotheses to use the initial examples $\mathcal{S}_{\mathrm{init}}$ for initializing rewards.
By allowing the hypotheses to initialize reward with more examples, the accuracy lies closer to its true value, allowing for fair comparison between earlier generated hypotheses and newer ones. 

\paragraph{Dynamic hypotheses update} In \cref{alg:hyp_gen}, we display how we generate and update the hypotheses pool $\mathcal{H}$. In particular, we add an example $s$ to the wrong example bank $\mathcal{W}$ if the number of hypotheses that incorrectly predict $s$ is greater than $w_{hyp}$. In our implementation, we use a linearly increasing $w_{hyp}$ as training time $t$ increases. This allows our algorithm to update the hypotheses more frequently at early stage of training, and less frequently at the end.

\subsection{Inference method implementations}
\label{appendix:inference_details}
\paragraph{Filter and weighted vote} In order to filter the hypotheses, we iterate through the top $k$ hypotheses ranked by reward. 
For each hypothesis, we ask the Large Language Model (LLM) if it is relevant.
Thereafter, for each of the relevant hypotheses, the LLM is prompted to use the hypothesis to make predictions. 
Then, for each predicted label, we add up the accuracy scores from the hypotheses that outputted that particular label.
The final label is the one that has highest total accuracy score.

\paragraph{One-step adaptive and two-step adaptive inference} The detailed framework of our adaptive inference methods is split into two parts - hypotheses pruning and hypotheses selection. In the case where we have a large number of hypotheses, it is likely that some hypotheses in $\mathcal{H}$ have overlaps or are paraphrases of each other. 

We address this issue with the following procedure:
\begin{enumerate}
    \item During training, we record the examples that each hypothesis correctly predicts.
    \item Then we create one-hot encodings for each hypothesis, where the $i$-th element of the one-hot encoding is 1 if the hypothesis correctly predicts the $i$-th example, and 0 otherwise. We subsequently compute a similarity matrix between each pair of hypotheses by taking the pairwise cosine similarities.
    \item Lastly, we create a linear program with the objective of maximizing the sum of accuracies of the selected hypotheses, subject to the constraint that every pair of the selected hypotheses has a similarity score below a predefined threshold $\gamma$.
\end{enumerate}
After pruning the set of hypotheses, we prompt the LLM to pick one hypothesis for its final prediction, as described in \cref{sec:hypothesis-based-inference}. 
For the single-step adaptive inference, we ask the LLM to select a hypothesis and make a prediction in one prompt. On the other hand, with the two-step adaptive inference, we first prompt the LLM to select a hypothesis and then prompt the LLM again to make a prediction based on the selected hypothesis. 

\subsection{Hyperparameters}
\label{appendix:hyperparams}
For the training stage, we set a limit on the hypothesis bank size, experimenting with sizes $H=3$  and $H=20$ to determine the impact of utilizing a larger number of hypotheses. Throughout all the experiments, we use the reward coefficient $\alpha=0.5$, $w_{max}=10$, $\mathtt{num\_init}=10$, and we have two different sets of the rest of hyperparameters for hypothesis bank sizes of $3$ and $20$. 
\begin{itemize}
    \item With $H=3$, we use $k=2$ and generate $1$ hypothesis per update. For inference, we employ all $3$ hypotheses for filter and weighted vote. For single-step and two-step adaptive inference, we use all $3$ hypotheses with $\gamma=0.3$ and provide $5$ examples to each hypothesis.
    \item In the case of $H=20$, we use $k=10$ and generate $5$ hypotheses per update. Then we take the top $5$ hypotheses, ranked by their training accuracies, for filter and weighted vote. For single-step and two-step adaptive inference, we use the top $5$ hypotheses with $\gamma=0.7$ and provide $5$ examples each.
\end{itemize}

\subsection{Licensing Details}
\label{appendix:license}

The \deceptive and \retweet datasets have not been released with any licenses, but are free to use for research purposes based upon the authors. 
The \headline dataset is released under the Creative Commons Attribution 4.0 International License. 
The \shoe dataset will be released under the same licensing as this work, CC BY 4.0 License, should it be accepted.

In regards to models, we find that \gpt and \claude are all proprietary models and are not released under any open-source licenses.
On the other hand, \mixtral is released under the Apache License 2.0. \roberta is not released under specific licensing but is free to use for research purposes. 
However, \llama is released under their own licensing found at \url{https://ai.meta.com/llama/license/}.

Per our extensive search, we find that we are in compliance with the licensing agreements of all the datasets and models used in this work.

\section{Detailed Results}
\label{appendix:detailed-results}

\subsection{\ours Performance across inference strategies}
\label{sec:performance_inference}
\begin{figure*}[t]
\centering
\begin{subfigure}[b]{0.45\textwidth}
\includegraphics[width=\textwidth]{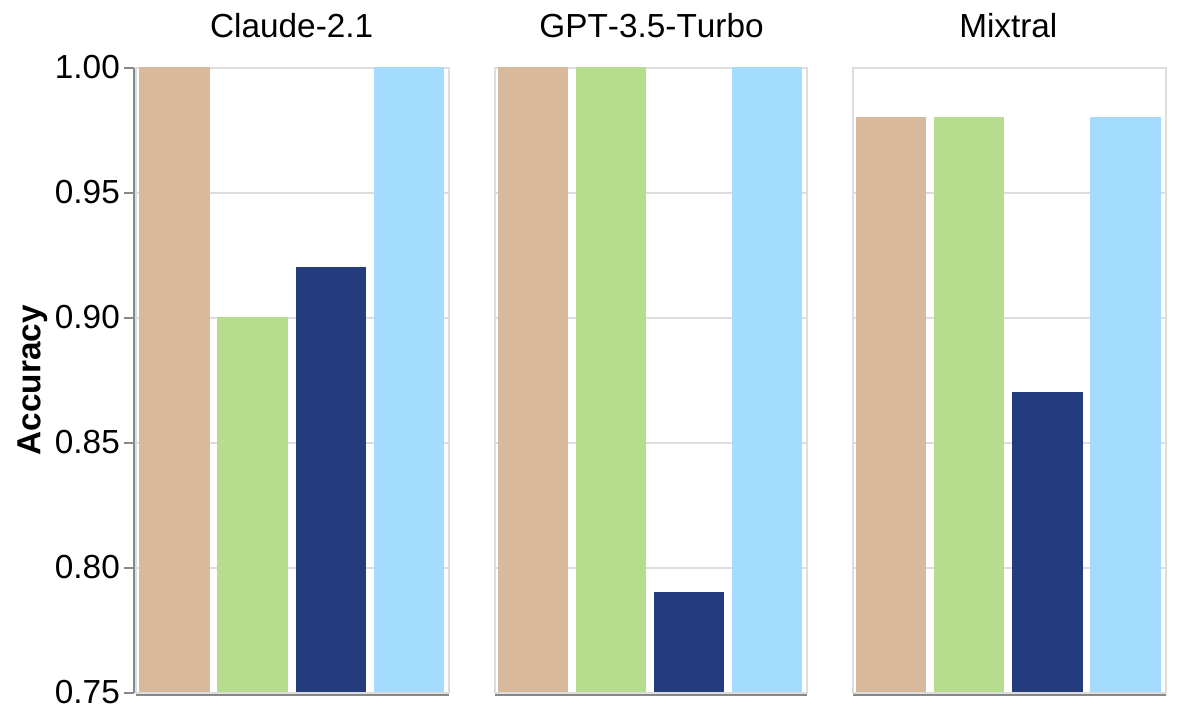}
\label{fig:inference_strategies_shoe}
\vspace{-4.1mm}
\caption{\shoe}
\end{subfigure}
\hspace{0.05\textwidth}
\begin{subfigure}[b]{0.45\textwidth}
\includegraphics[width=\textwidth]{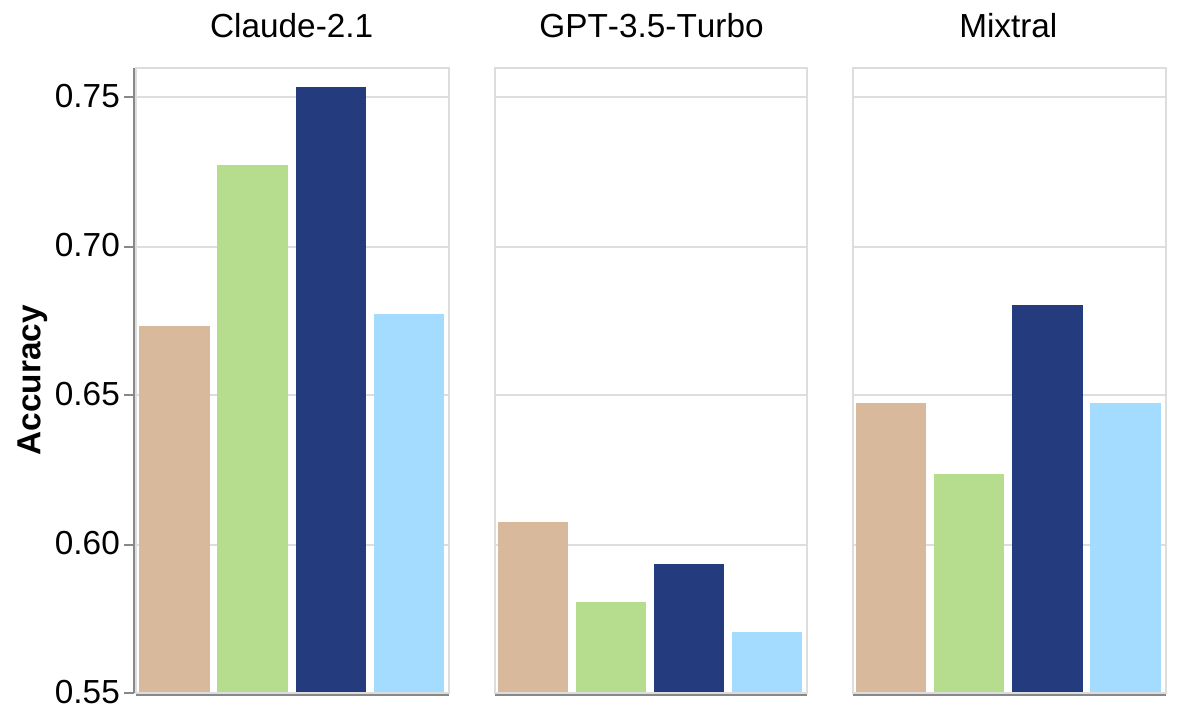}
\caption{\deceptive}
\end{subfigure}
\begin{subfigure}[b]{0.45\textwidth}
\includegraphics[width=\textwidth]{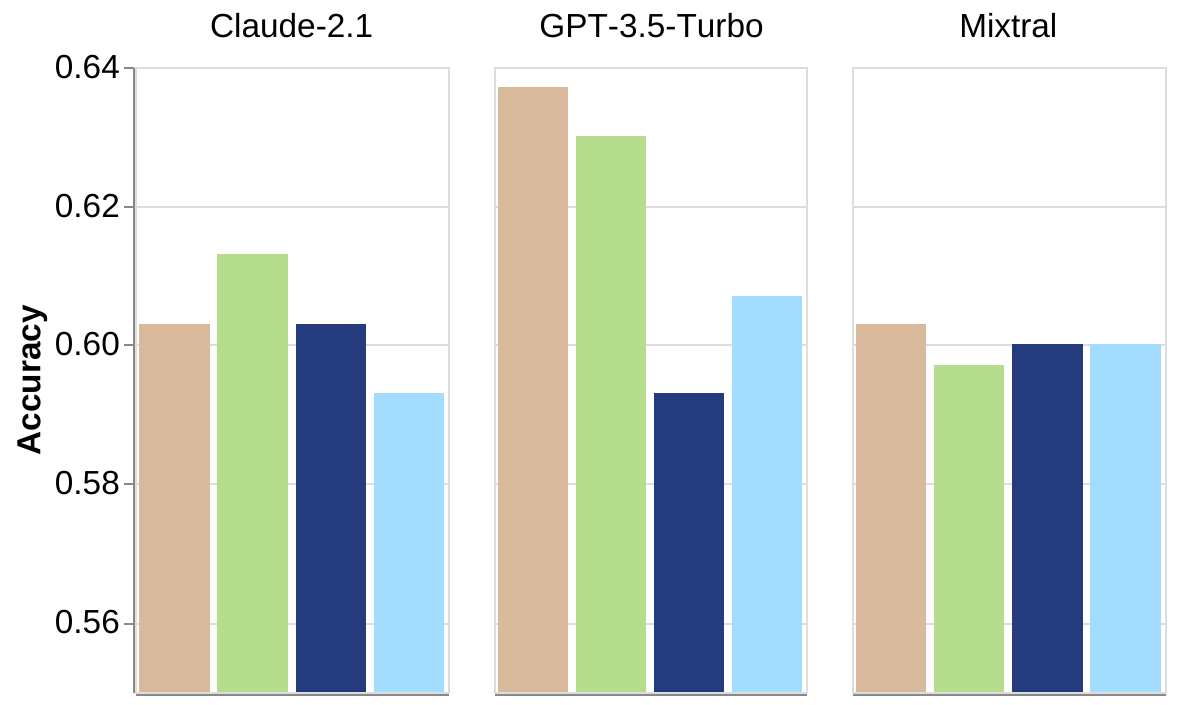}
\caption{\headline}
\end{subfigure}
\hspace{0.05\textwidth}
\begin{subfigure}[b]{0.45\textwidth}
\includegraphics[width=\textwidth]{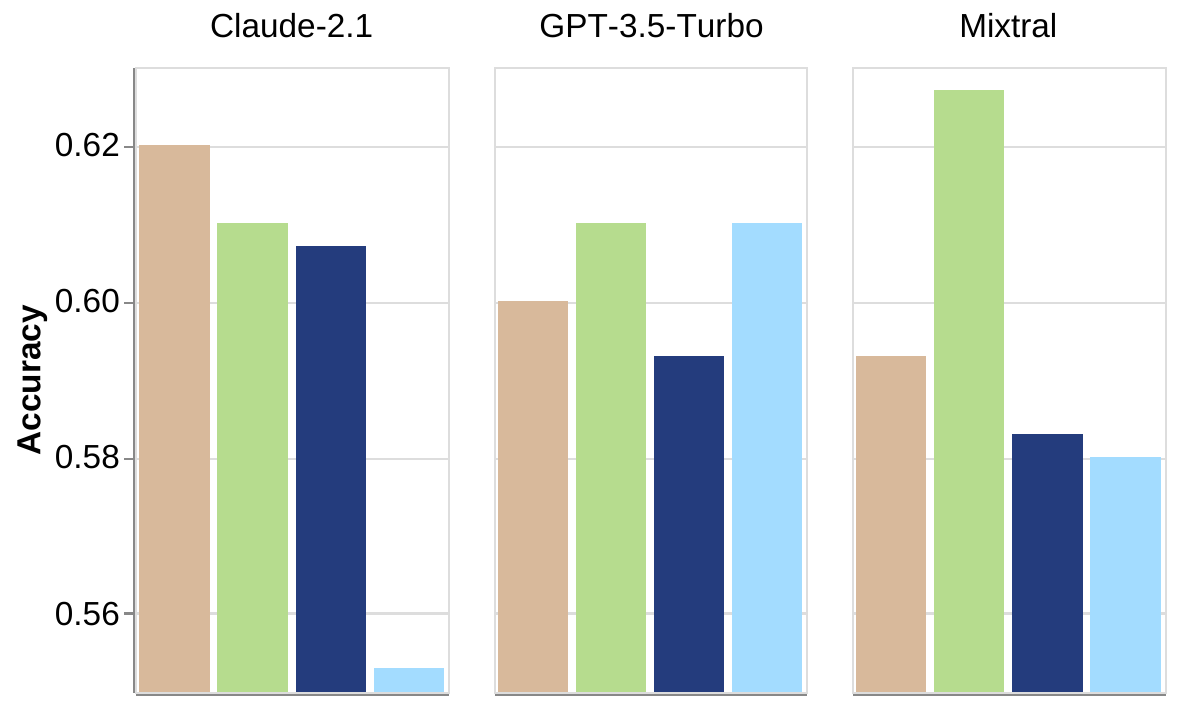}
\caption{\retweet}
\end{subfigure}
\caption{
    \ours results with different inference strategies. 
    Best-accuracy hypothesis is sufficient for getting good performance on \shoe and \headline. 
    Single-step adaptive hypothesis-based inference is the most effective on \deceptive.
    Filter and weighted vote is best on \retweet.
    }
\label{fig:inference_strategies}
\end{figure*}

\cref{fig:inference_strategies} presents the best results for all of our inference strategies, considering every dataset and all hyperparameter configurations.

For \shoe, we observe that all the models perform effectively by using the best hypothesis inference strategy.
Surprisingly, \mixtral is unable to perform perfectly. 
This is because despite generating the hypothesis that fully describes the data, \mixtral opts not to apply the hypotheses, favoring to choose a random label for the sake of ``variety''. 
Both \gpt and \mixtral display similar patterns across the inference strategies, with best-accuracy hypothesis, filter and weighted vote, and two-step adaptive inference all having comparable performance. 
However, for all models we find single-step adaptive inference drops in accuracy. 
Given that two-step adaptive inference performs well, it is likely that the long prompt causes the model difficulty in choosing the correct hypotheses. 
For \claude, we see that filter and weighted vote drops in performance. 
As this method searches for relevant hypotheses, the model is likely finding that inaccurate patterns relevant, which end up outweighing the inference of the best hypothesis. 

For \deceptive, \claude is the best performing model across all inference policies. 
Across the models, we highlight that single-step adaptive inference method works best for this dataset. 
In this inference method, the prompt specifically includes the aims of determining if a review is deceptive. This likely helps the model use the context provided to better decide which set of example resembles the test example most.
Hence, splitting up the prompt may have caused performance to suffer. 

We find that \headline is the most challenging dataset. 
As mentioned in \cref{sec:datasets}, the original dataset was created with both images and headlines paired together. 
In our version of the dataset, we only use the headlines, so we are missing a crucial variable that contributes to understanding click behavior.
Therefore, based off only headlines, it is difficult to generate hypotheses that truly capture the data.
Despite this challenge, we note that our hypotheses can still adeptly capture a large portion of data with 63.7\% being our highest accuracy.
Specifically, we find that the best-accuracy hypothesis strategy performs best. 
We also note that filter and weighted vote can provide strong performance as in the case of \claude and \gpt, suggesting that hypotheses corroborating with each other can lead to better performance. 
We observe that \gpt is the best performing model here, with all inference policies (aside from single-step adaptive) having high accuracy. 

Finally, over the \retweet dataset, we find that the filter and weighted vote is the best choice for inference policy, with it being the best inference method for \gpt and \mixtral.
This indicates that using hypotheses in conjunction is useful as multiple variables together adeptly characterize the dataset.
The performance of the rest of the inference policies has no clear pattern over this dataset. 

We also present our results with confidence intervals. We specifically see that compared to the Oracle Methods, \ours shows performance statistically significant benefits when comparing to the 200 training examples for \headline and \retweet. However, this is not the case for \deceptive, because there are word level features that make the task easier for unsupervised methods. We note that \ours has statistically significant performance increases for \deceptive with \claude and \mixtral and for \retweet with \claude and \mixtral. 
\begin{table*}[t]
    \centering
    \resizebox{\textwidth}{!}{%
    \tiny
    \begin{tabular}{@{}llrrrr@{}}
        \toprule
                           &                    & \textsc{Shoe}  & \textsc{Deceptive} & \textsc{Headline}   & \textsc{Tweet}        \\ 
        Models             & Methods            & \textsc{Sales} & \textsc{Reviews}   & \textsc{Popularity} & \textsc{Popularity}   \\ \midrule \midrule

        \roberta (Oracle) & Train 200            & 100.0 $\pm$ 0.0      & 84.0  $\pm$ 4.2         & 49.0  $\pm$ 5.7     & 50.7  $\pm$ 5.7    \\
                         & Train 1000           & 100.0   $\pm$ 0.0   & 91.0   $\pm$ 3.2         & 60.0   $\pm$ 5.5    & 62.0  $\pm$ 5.5    \\ \midrule
        \llama (Oracle)      & Train 200            & 100.0   $\pm$ 0.0   & 88.7    $\pm$ 3.6          & 49.7 $\pm$ 5.7    & 50.3 $\pm$ 5.7  \\ 
                          & Train 1000   & 100.0 $\pm$ 0.0       & 92.3   $\pm$ 3.0           & 60.0 $\pm$ 5.5    & 51.3 $\pm$ 5.7  \\ \midrule

		\claude		 & Few shot				 & 75.0	$\pm$ 4.9	 & 51.0	$\pm$ 5.7			 & 60.0 $\pm$ 5.5		 & 0.3$^*$	$\pm$ 0.6	 \\ 
						 & \ours			 & {\bf 100.0}	$\pm$ \textbf{0.0}	 & {\bf 75.3} $\pm$ \textbf{4.9}				 & {\bf 61.3} $\pm$ \textbf{ 5.5}		 & {\bf 62.0} $\pm$ \textbf{5.5}		 \\ \midrule 
		\mixtral	& Few shot				 & 79.0	$\pm$ 4.6	 & 56.3	$\pm$ 5.6			 & 55.3	$\pm$ 5.6	 & 48.7	$\pm$ 5.7	 \\ 
						 & \ours			 & {\bf 98.0}	$\pm$ \textbf{1.6}	 & {\bf 68.0} $\pm$ \textbf{5.3}			 & {\bf 60.3}	$\pm$ \textbf{5.5}	 & {\bf 62.7}	$\pm$ \textbf{5.5}	 \\ \midrule 
		\gpt	    & Few shot				 & 49.0 $\pm$ 5.7		 & 55.0	$\pm$ 5.6			 & 60.0		$\pm$ 5.5 & {\bf 62.0}$\pm$ \textbf{5.5}		 \\ 
						 & \ours			 & {\bf 100.0}	$\pm$ \textbf{0.0}	 & {\bf 60.7}	$\pm$ \textbf{5.5}			 & {\bf 63.7}	$\pm$ \textbf{5.4}	 & 61.0	$\pm$ 5.5	 \\ \bottomrule
    \end{tabular}
    }
    \caption{
        Table with 95\% confidence interval for Few shot results and \ours for our best results.
    }
    \label{tab:performance_with_CI}
\end{table*}

\subsection{\ours Performance across training examples}
\label{sec:performace_training}
\begin{figure*}[t]
    \centering
    \includegraphics[width=\textwidth]{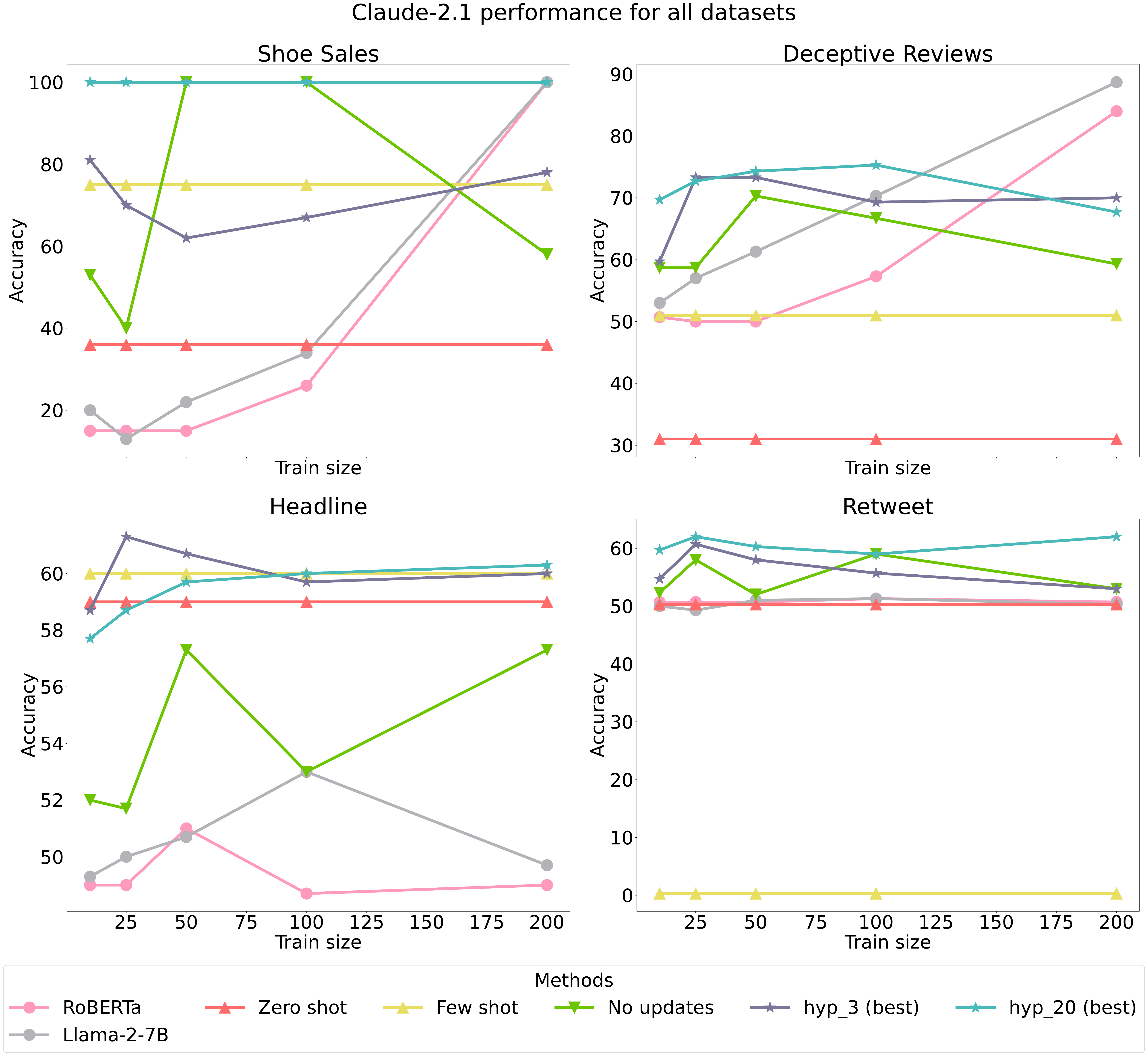}
    \caption{
        \claude results for baselines, \ours (no update), and \ours (best) with hypothesis bank size 3 and 20 across multiple training samples
    }
    \label{fig:claude_line_chart}
\end{figure*}

\cref{fig:claude_line_chart} presents the results for the performance of \ours with \claude as the training examples change. 
We observe that for all of our datasets, \ours outperforms zero-shot and few-shot learning generally for all training examples in \shoe and \retweet. 
In \headline, we find that the model needs to use 200 examples to outperform them. 
We highlight that \ours outperforms the No Updates method for all training examples across the four datasets when using a hypothesis bank size of 20. 
When using a hypothesis bank size of 3, we find that in \retweet, \ours is able to outperform the No Updates method, but is unable to as the training examples increase.
In \shoe we observe that it is largely worse because we set $k$ (as discussed in \cref{sec:hypotheses-generation}) to be 1, which causes difficulty in finding the best hypothesis.
It is unclear what the optimal number of training examples is across the datasets, as using more examples does not necessarily increase accuracy.
\begin{figure*}[h]
    \centering
    \includegraphics[width=\textwidth]{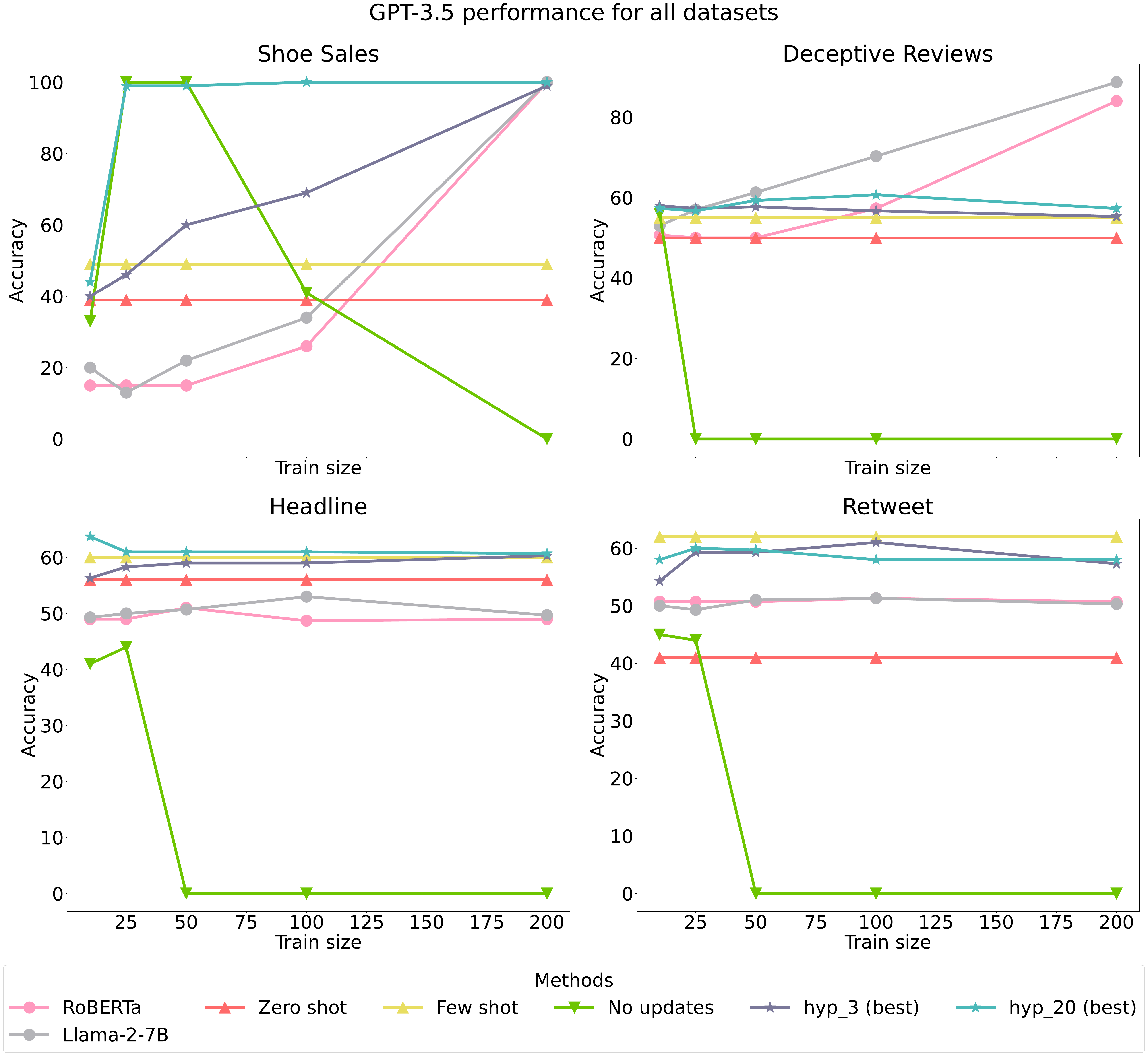}
    \caption{
        \gpt results for baselines, \ours (no update), and \ours (best) with hypothesis bank size 3 and 20 across multiple training samples
    }
    \label{fig:gpt_line_chart}
\end{figure*}

\cref{fig:gpt_line_chart} displays the accuracy for \ours with \gpt for the different training examples.
We observe that unlike \ours performance with \claude, our results are mixed for when our method outperforms the few shot inference. 
Specifically, in \retweet, the few shot inference surpasses our results, indicating that in this set hypotheses provide less benefits than using examples.
As \ours exceeds the accuracy of zero shot's, the proposed method still provides benefits to the base model.
Similar to the results on \claude, we outperform \roberta and \llama on all datasets aside on \deceptive for all training examples. 
\ours surpasses the performance of the No Update strategy generally for all training examples. 
We note that due to the limited context window of \gpt, the No Update strategy fails as it is unable to accept training examples. 
\ours effectively bypasses this issue by iteratively going through test examples, as opposed to feeding them into the model all at once. 

\begin{figure*}[t]
    \centering
    \includegraphics[width=\textwidth]{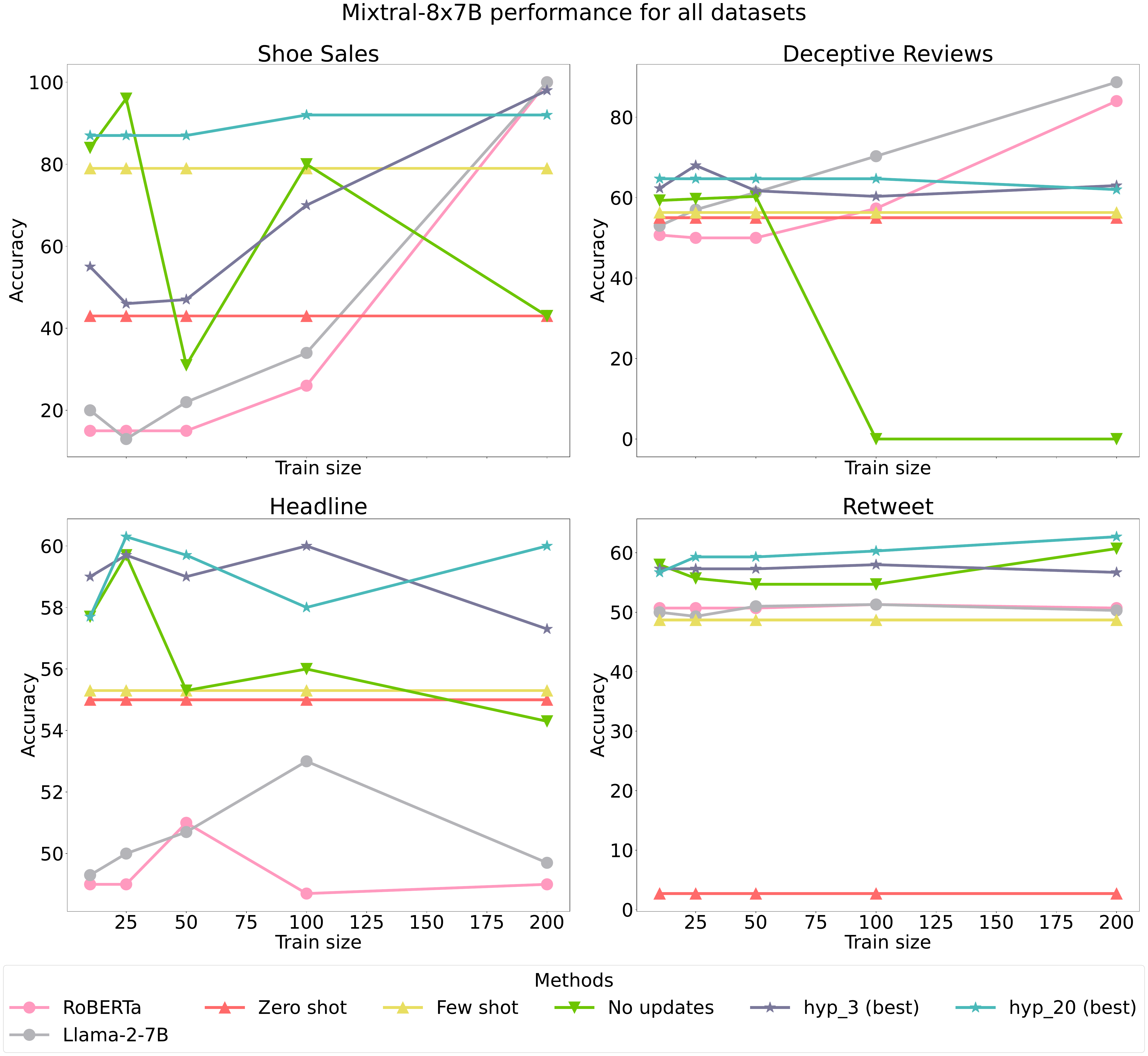}
    \caption{
        \mixtral results for baselines, \ours (no update), and \ours (best) with hypothesis bank size 3 and 20 across multiple training samples
    }
    \label{fig:mixtral_line_chart}
\end{figure*}

In, \cref{fig:mixtral_line_chart}, the performance of \ours for varying training examples with \mixtral is shown. 
\ours outperforms the zero shot and few shot strategies for all datasets, aside from \shoe, where the proposed method requires 200 examples to outperform few shot learning.
Similary, we note that \ours surpasses the performance of \roberta and \llama for \headline, \retweet, and generally for \shoe. 
As mentioned in \cref{sec:performance_inference}, despite \mixtral finding the best hypothesis, it occasionally refuses to choose the correct label to encourage ``variety'', which causes \roberta and \llama to outpeform \ours.
In comparison to the No Update results, we find that in \deceptive and \headline, \ours matches or exceeds this method. 
For \shoe, we find that with hypothesis bank 3, \ours must use 200 examples, to finally converge to the correct hypothesis. 
On the other hand, for \retweet, No Update surpasses the \ours with hypothesis bank size 3 after using 200 training examples.
This may occur as using 3 hypotheses is too limited to adeptly describe the dataset, causing accuracy to suffer.

\subsection{Full OOD results}
\label{sec:full_ood_results}

\begin{table*}[t]
    \centering
    \resizebox{\textwidth}{!}{%
    \small
    \begin{tabular}{@{}llcccc@{}}
        \toprule
        Models           & Methods               	 & \textsc{IND Deceptive Reviews} & \textsc{OOD Deceptive Reviews} \\ \midrule \midrule
        RoBERTa (Oracle) & Train 200				 & 84.0 & 73.0 (\decrease$11.0$) \\
                         & Train 1000				& 91.0 & 79.7 (\decrease$11.3$) \\\midrule 
        Llama-2-7B (Oracle) & Train 200				 & 88.7 & 78.7 (\decrease$10.0$) \\
                         & Train 1000				& 92.3 & 88.7 (\decrease$3.6$) \\\midrule                  
        \claude			 & Zero shot				 & 31.0 & 27.7 (\decrease$3.3$) \\ 
                         & Few shot					& 51.0 & 41.7 (\decrease$9.3$) \\ 
                         & \ours (Best-accuracy hypothesis)			 & 67.3 & 71.7 (\increase$4.4$) \\ 
                         & \ours (Filter and weighted vote)			 & 68.0 & 74.7 (\increase$6.7$) \\ 
                         & \ours (One-step adaptive) & 70.0 & 68.3 (\decrease$1.7$) \\ 
                         & \ours (Two-step adaptive) & 67.7 & 70.7 (\increase$3.0$) \\ \midrule 
        \mixtral		 & Zero shot				 & 55.0 & 49.7 (\decrease$5.3$) \\ 
                         & Few shot					& 56.3 & 49.0 (\decrease$7.3$) \\ 
                         & \ours (Best-accuracy hypothesis)			 & 61.3 & 64.7 (\increase$3.4$) \\ 
                         & \ours (Filter and weighted vote)			 & 62.0 & 61.0 (\decrease$1.0$) \\ 
                         & \ours (One-step adaptive) & 63.0 & 54.7 (\decrease$8.3$) \\ 
                         & \ours (Two-step adaptive) & 61.3 & 64.7 (\increase$3.4$) \\ \midrule 
        \gpt			 & Zero shot				 & 50.0 & 49.0 (\decrease$1.0$) \\ 
                         & Few shot					& 55.0 & 52.0 (\decrease$3.0$) \\ 
                         & \ours (Best-accuracy hypothesis)			 & 57.3 & 60.7 (\increase$3.4$) \\ 
                         & \ours (Filter and weighted vote)			 & 55.3 & 55.7 (\increase$0.4$) \\ 
                         & \ours (One-step adaptive) & 55.7 & 51.7 (\decrease$4.0$) \\ 
                         & \ours (Two-step adaptive) & 54.7 & 59.0 (\increase$4.3$) \\ \bottomrule
        \end{tabular}
        }
        \caption{
            Performance of baselines and compared to our methods on the out-of-distribution deceptive reviews and \deceptive.
        }
        \label{tab:ood_results_full}
    \end{table*}

\cref{tab:ood_results_full} shows results for the OOD deceptive reviews dataset for all inference strategies for each model. 

We find that \ours outperforms both zero shot and few shot learning across all models and inference policies. 
The best-accuracy hypothesis and two-step adaptive inference methods are the most robust, showing an average increase of 3.7\% and 3.6\% respectively.
We claim that although the filter and weighted vote strategy at first glance may seem to have mixed performance, the method is still robust. 
The drop in accuracy for \mixtral with filter and weighted is minimal (1\%), and both \gpt and \claude exhibit increases in accuracy.
Hence, the inference policy is consistent across \deceptive and the OOD deceptive review datset. 
Interestingly, the single-step adaptive inference method exhibits drops in performance despite being the best performing inference model in \deceptive.
In single-step adaptive inference, the LLM sees both the hypotheses with the sets of examples along with the final question of determining whether the review is deceptive. 
Even though the LLM is prompted to only use one chosen hypotheses, these training examples from \deceptive negatively impact the model because they are part of the context and are thus inherently used by LLMs. 
On the other hand, for two-step adaptive inference, since there is a dedicated prompt for hypothesis selection, the application of the hypothesis is unaffected from the \deceptive training samples.

\section{Qualitative Analysis on Generated Hypotheses}
\label{appendix:hypothesis-analysis}

We include findings from the generated hypotheses on \deceptive, \headline, and \retweet datasets in Table \ref{tab:hypotheses_analysis}. The table shows that the a good number of the hypotheses are supported by existing findings, while others are novel.
This suggests that the generated hypotheses are grounded in existing literature and can be used to guide future research.

\begin{table*}[t]
    \centering
    \renewcommand{\arraystretch}{1.25} %
    \resizebox{\textwidth}{!}{%
    \small
    \begin{tabular}{@{}lp{7cm}p{4.25cm}@{}}
    \toprule
    \textbf{Dataset} & \textbf{Finding} & \textbf{Supported/Novel} \\ \midrule \midrule
    \deceptive & Deceptive reviews contain more emotional terms. & \citet{li2014generalrule} \\ 
               & Deceptive reviews are more likely to use superlatives. & \citet{ott2011finding} \\
               & Deceptive reviews contain hearsay or information that could not have been directly experienced. & \citet{ott2011finding} \\
               & Deceptive reviews tend to be more exaggerated. & \citet{anderson2014reviews} \\
               & Truthful reviews tend to use more balanced and objective tone. & \citet{anderson2014reviews} \\
               & Truthful reviews could mention the reviewer's purpose for staying at the hotel (e.g., business trip, vacation). & Novel \\
               & Truthful reviews would mention weddings or special occasions. & Novel \\
               & Truthful reviews may contain information about reviewer's expectations and previous hotel experiences. & Novel \\
               & Truthful reviews would acknowledge the reviewer's personal biases or preferences. & Novel \\
               & Deceptive ones may present the reviewer's opinion as objective facts. & Novel \\
               & Truthful reviews may contain reviewers' past experiences or future travel plans. & Novel \\
    \midrule
    \headline  & Concreteness helps. & \citet{sadoski2000engaging} \\ 
               & Revealing something new helps. & \citet{banerjee2023language} \\
               & Using vivid language and imagery helps. & \citet{banerjee2023language} \\
               & Headlines with high intensity of emotions would be clicked more. & \citet{banerjee2023language} \\
               & Action-oriented headlines are clicked more. & \citet{banerjee2023language} \\
               & Humorous headlines are clicked more. & Novel \\
               & Controversial headlines are clicked more. & Novel \\
               & Headlines that frame the content in a personal or relatable way are clicked more. & Novel \\
    \midrule
    \retweet   & Short and concise tweets are retweeted more. & \citet{gligoric2019causal} \\ 
               & Tweets with emotional tones are retweeted more. & \citet{chenhao2014retweet} \\

               & Including specific details (e.g., dates, locations) are associated with more retweets. & Novel \\
               & Including statistics and data are associated with more retweets. & Novel \\
               & Mentioning influential individuals or organizations leads to more retweets. & Novel \\
               & Including links to additional content (e.g., articles, videos) leads to more retweets. & Novel \\
               & Tweets with a call to action or urgency are found to be retweeted more. & Novel \\
    \bottomrule
    \end{tabular}
    }
    \caption{Summary of generated hypotheses (on the real-world datasets) and whether they support existing findings or are novel.}
    \label{tab:hypotheses_analysis}
    \end{table*}

\end{document}